\documentclass[a4paper,conference,left=0.514in,right0.514in,top=0.75in,bottom=1.569in]{IEEEtran}

\usepackage{microtype}
\usepackage{graphicx}
\usepackage{subfigure}
\usepackage{epsfig}
\usepackage{booktabs} 
\usepackage{hyperref}
\usepackage{amsmath}

\begin{document}

\title{Filtered Batch Normalization}

\author{\IEEEauthorblockN{András Horváth, Jalal Al-afandi}
\IEEEauthorblockA{Peter Pazmany Catholic University\\
Faculty Of Information Technology and Bionics\\
Budapest, Hungary\\
Email: horvath.andras@itk.ppke.hu,\\ alafandi.mohammad.jalal@itk.ppke.hu}
}

\maketitle

\begin{abstract}
It is a common assumption that the activation of different layers in neural networks follow Gaussian distribution. This distribution can be transformed using normalization techniques, such as batch-normalization, increasing convergence speed and improving accuracy. 
In this paper we would like to demonstrate, that activations do not necessarily follow Gaussian distribution in all layers. Neurons in deeper layers are more selective and specific which can result extremely large, out-of-distribution activations.

We will demonstrate that one can create more consistent mean and variance values for batch normalization during training by filtering out these activations which can further improve convergence speed and yield higher validation accuracy.
\end{abstract}
\IEEEpeerreviewmaketitle

\section{Introduction}

The application of normalization methods is ubiquitous in signal processing and has a long history in machine learning.
Various techniques were introduced in the past decade, such as local response normalization, which was one of the cornerstone components of AlexNet \cite{krizhevsky2012imagenet}, instance normalization \cite{ulyanov2016instance}, which is commonly applied in style transfer networks or layer normalization \cite{ba2016layer}, which is mostly applied in recurrent neural networks. The most generally applied and in most cases best performing normalization technique still remained batch normalization (BN), which was introduced in \cite{ioffe2015batch}.


Batch normalization became an important building block of neural networks in the past five years. It was demonstrated in various tasks that this method can accelerate network training and results higher test accuracy in practice, if mini-batch size is sufficiently high.

The beneficial effect of batch normalization was introduced in \cite{ioffe2015batch} hypothesizing that the reduction of internal covariate shift -which is the imposed change in the input distribution of layers triggered by the updates of the preceding layers - can result faster convergence.

Although \cite{santurkar2018does} demonstrated that mitigating internal covariate shift plays only a minor role in the effectiveness of BN, it is still beneficial as a regularizer which reparametrizes the activations during training to make the optimization problem more stable by creating a smoother loss landscape and increasing the predictability of gradients rendering a robustness against exploding or vanishing gradients, hyperparameters and initialization sensitivity.

BN is useful in training, but can be difficult to be applied during inference, since usually a single input is presented instead of mini-batches. Batch re-normalization \cite{ioffe2017batch} was introduced to mitigate this problem, where instead of the calculated first and second order moments, smoothed moving averaged mean and variance values are used. The same paper also suggested that mean and variance values can be inconsistent in case of smaller mini-batches, but smoothing them with moving averages can help to result more consistent values for normalization.

In our paper we will demonstrate that BN with or without running averages 
can still result inconsistent mean and variance values even with fairly large mini-batches (such as 128 or 256) and this is caused by unlikely large activations in the network (assuming that they follow Gaussian distribution).
If running averages are not used in training the output can heavily depend on the last training batch, since this will determine the results, meanwhile with smoothing this difference is decreased, but remains for more iterations.
We will demonstrate that the specificity of network activations for certain features works against the common assumption that activations can be described well by Gaussian distributions. Exploiting these facts, we will demonstrate a method which filters out these out-of-distribution samples in batch normalization using robust statistics, resulting more consistent moments and faster convergence.

Our method adds an additional mean and variance calculation step in training, which does not result a significant increase in the number of operations in case of complex networks. Also our method has no affect on the number of additional operations in inference, hence the trained networks can be executed with the same computational performance as in the case of BN.

It was also recently demonstrated that although batch normalization is still the best performing approach in case of sufficiently large mini-batches, the performance can deteriorate in case of smaller ones. Unfortunately, in case of complex models like Resnet-101 \cite{he2016deep} or Mask-RCNN \cite{he2017mask} only small mini-batches fit in the GPU memory (in case of Mask-RCNN  with ResNext-101\cite{xie2017aggregated} backbone, only mini-batches of two images can fit the memory limit of an NVIDIA RTX 2080 TI), resulting inferior performance.
This can be improved using group normalization \cite{wu2018group} where the moments for normalization are calculated over multiple channels and spatial positions, but not over instances of the mini-batch. We will demonstrate that our method can improve group normalization as well. In theory, our approach -the robust calculation of mean and variance values- could be used in other methods as well (such as layer normalization and filter response normalization \cite{singh2019filter}), but the investigation of these possibilities is out of the scope of the current paper.

Our paper is organized as follows.
In Section \ref{SecActivations} we will first briefly introduce traditional batch normalization and demonstrate the caveats of the general assumption, that activations in deep neural network follow Gaussian distribution.
In Section \ref{SecFilteredBatchNorm} we will introduce filtered batch normalization: a novel approach which calculates robust statistics and results faster convergence.
In Section \ref{SecResults} we will compare our method to traditional batch normalization on commonly applied datasets (MNIST \cite{lecun1998mnist}, CIFAR-10 \cite{krizhevsky2014cifar}, ImageNet 2012 \cite{deng2012imagenet}, MS-COCO \cite{lin2014microsoft}) and network architectures (LeNet-5 \cite{lecun1998gradient}, AlexNet \cite{krizhevsky2012imagenet}, Mask-RCNN \cite{he2017mask} with ResNext-101 backbone \cite{xie2017aggregated}, VGG \cite{simonyan2014very} and ResNet-50 \cite{he2016deep}).
In Section \ref{SecConclusion} we conclude our findings.

\section{Batch Normalization and The Distribution of Neural Network Activations}\label{SecActivations}

\subsection{Batch normalization}

The aim of batch normalization is to result coherent output distributions in every iteration at each layer. Assuming that network activations follow a Gaussian distribution, which can be fully described by the mean and variance values, we can transform the output distribution of a network to zero mean and variance of one. After calculating the first two moments (mean and variance), batch normalization scales and shifts the activations using trainable parameters maintaining landscape flexibility. We also have to note that because of this scaling the exact mean and variance which are selected for normalization does not matter from an algorithmic point of view, the moments just have to be constant to ensure the same output distributions.

Based on this, batch normalization can be described by the following formula:

\begin{equation}\label{EqBatchNorm}
    y_i = \gamma \frac{(x_i - \mu_i)}{\sigma_i}  +  \beta
\end{equation}
Where $x_i$ are the activations of layer or kernel $i$, $y_i$ are the transformed activations, $\gamma$ and $\beta$ are trainable parameters and $\mu_i$ and $\sigma_i$ are the mean and the standard deviation, which are calculated by the following equations:

\begin{equation}\label{EQMean}
    \mu_i = \frac{1}{m}  \sum_{k \in S_i} x_k
\end{equation}

\begin{equation}\label{EQVar}
\sigma_i = \sqrt{\frac{1}{m}  \sum_{k \in S_i} (x_k - \mu_i)^2 + \epsilon } 
\end{equation}

Where $m$ is the number of neurons in the selected layer or kernel, $\epsilon$ is a small numerical value to avoid division by zero and  $S_i$ is the set of activations which are selected for normalization (this set of activations are determined differently in case of group, instance or layer normalization \cite{wu2018group}).

\subsection{Distribution of Neural Network Activations}

It is a common assumption that activations follow Gaussian distribution in neural networks before the application of the non-linear transfer functions.

This is true for untrained networks with randomly initialized weights following Gaussian distribution, but is also commonly hypothesized to be true for networks after or during training as well.

Although this assumption is helpful and led to many normalization techniques, it contradicts the assumption that neurons or kernels in convolutional neural networks (CNNs) are feature or class specific responding and emitting high activations only to certain features.
This specificity is obvious at the logit layer:  in  case of classification problems (but similarly in detection and segmentation as well) one expects a single neuron with large activation, which belongs to the specific class, meanwhile all other neurons should have minimal activations. This more Bernoulli-like distribution is compelled on the neurons during training.

The specificity of deeper kernels was demonstrated in visualization of network responses \cite{olah2018building} and attribute maps \cite{ancona2017towards} which showed that kernels in deeper layers of CNNs are feature specific and output high activations only if the proper output class is presented.

To demonstrate this we have measured the normalized activations in each layer in pretrained networks. We have selected VGG-16 \cite{simonyan2014very} with batch normalization as a reference to investigate the activations in each layer after batch normalization on the validation set of ImageNet 2012 \cite{deng2012imagenet}. The weights of the pretrained network were taken from the torchivision.models module\footnote{The weights can be downloaded from:
\href{https://download.pytorch.org/models/vgg16_bn-6c64b313.pth}{ \url{https://download.pytorch.org/models/vgg16_bn-6c64b313.pth}}} to ensure reproducability.

\begin{figure}[!htp]
\vskip 0.2in
\begin{center}
\centerline{\includegraphics[width=\columnwidth]{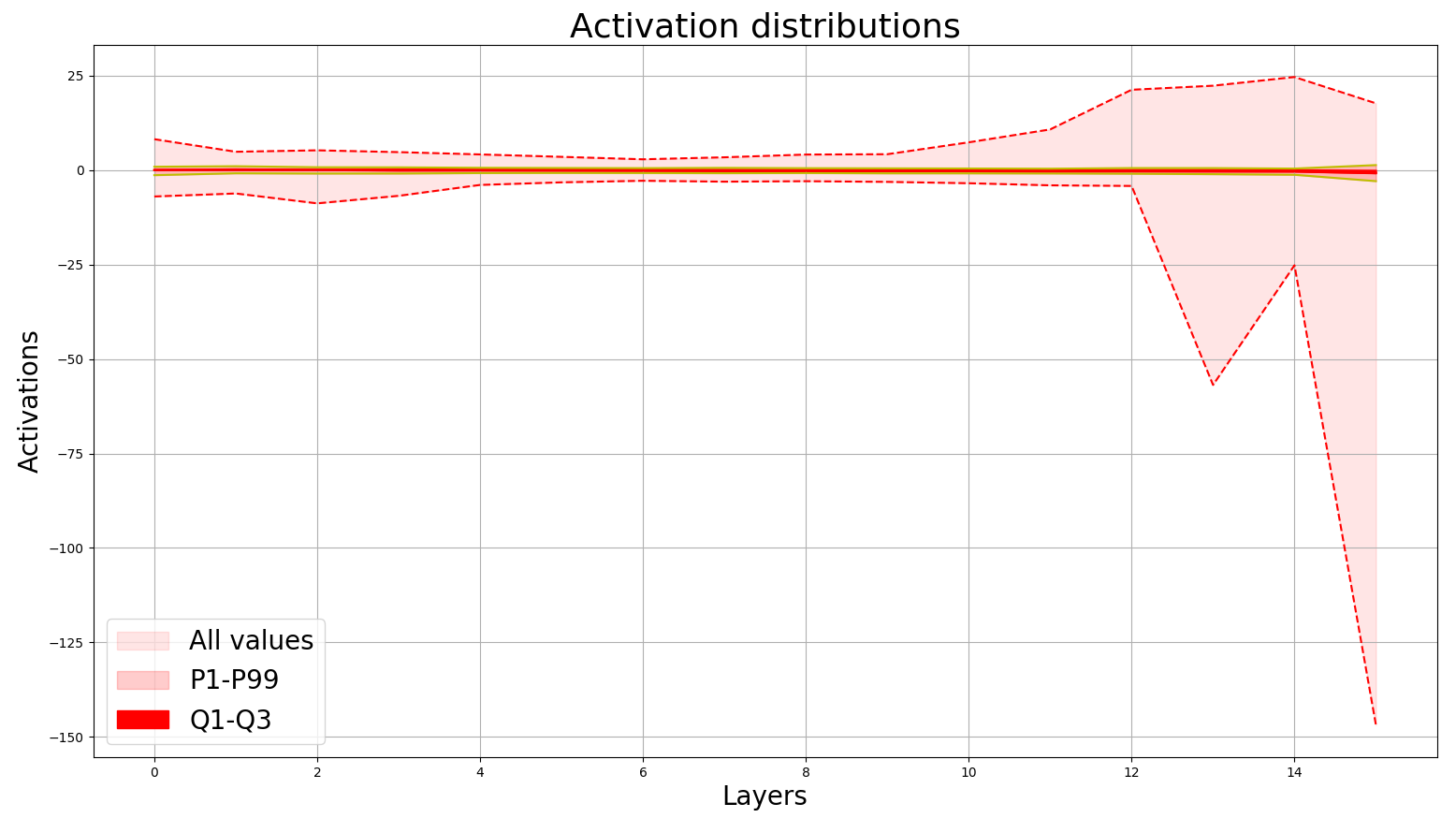}}
\caption{This figure depicts the distribution of the activations in a pretrained version of the VGG-16-BN architecture. The dashed lines display the maximum and minimum values in each layer, the golden lines contain $98\%$ of the activations and  $50\%$ of them is in the solid red region. This demonstrates that although the data has zero mean and variance of one, it contains outliers especially in layers closer to the logit layer.}
\label{FigVggMaxminActivations}
\end{center}
\vskip -0.2in
\end{figure}

\begin{figure}[!htp]
\vskip 0.2in
 \centering
    \subfigure{
    \includegraphics[width=\columnwidth]{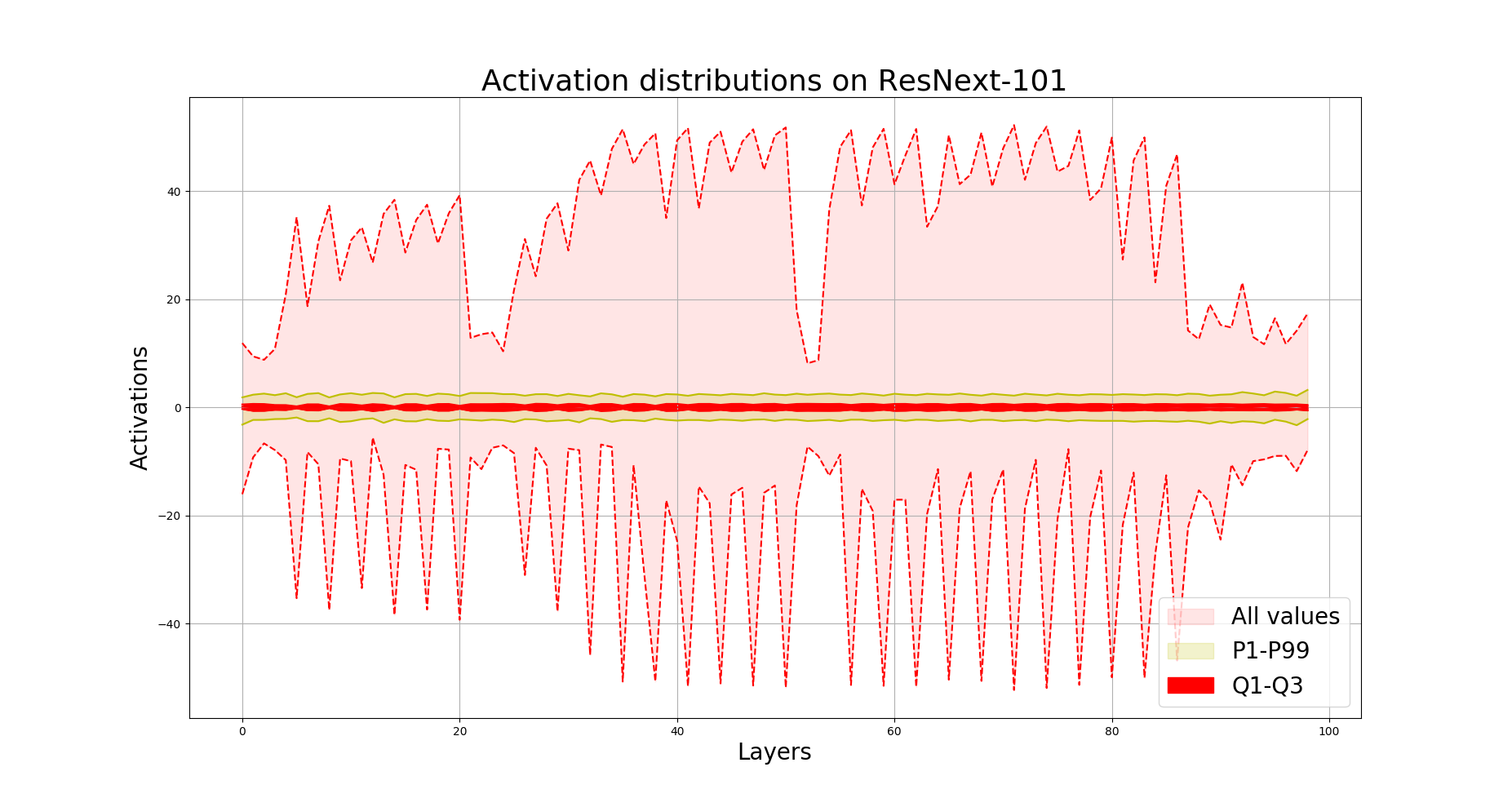}}
    \caption{
    This figure depicts the distribution of the activations in the ResNext-101 architecture after every batch normalization. As it can be seen the distribution is similar as in case of VGG-16. Although the distribution of the outliers is different (in VGG large activations have appeared at deep layers, here the distribution of the activations does not seem to be dependent on the layer depth. Also the appearance a periodicity can be seen in the minimal/maximal activations, which might depend on the position of the convolution in the bottleneck block of the residual structure- but wee have not investigated this in detail), it is still true that out-of-distribution activations appear in the network. We have to emphasize that on this plot only activations after convolutional layers are present. $98\%$ can be found in the golden band and the middle dark red band contains $75\%$ of the values. The pretrained weights of the model were downloaded from:
   \url{ https://download.pytorch.org/models/resnext101_32x8d-8ba56ff5.pth}}
\label{Figactivationsresnext101}
\end{figure}

We have used this network in inference mode, where previously learned constant values are used for normalization instead of calculating the current mean and variance. We investigated the magnitude of the activations in the batch normalization layer before parameters $\gamma$ and $\beta$ were applied, so all activations should have zero mean and variance of one. The activations according to the layers can be seen on Figure \ref{FigVggMaxminActivations}.
From this plot one can see that although $68\%$ of the activations are in a narrow band ($\pm\sigma$), which results a variance of one, there are some activations outside $\pm7\sigma$ in every layer. The three fully connected layers at the end of the network have especially extreme activations ($-140\sigma$), which should not happen in case of a Gaussian distribution.
These distributions were observed with other architectures as well such as in case of ResNext101 \cite{xie2017aggregated}, which can be seen in Fig. \ref{Figactivationsresnext101}. Although the distribution of extremely large activations is different, the presence of these values can still be observed (values outside the $40 \sigma$ band). The largest activations were observed at the last layer of the bottleneck blocks of the residual structure.

The presence of values outside the $7\sigma$ range should have a probability of 1/390,682,215,445. This means that considering a layer with 512 kernels and 16x16 positions, a value should appear out of every 2,980,668 input images, so even in the whole validation set of ImageNet the probability of the appearance of such extreme activations is as low as 0.0167. We have also investigated VGG-19 and ResNet-50 architectures and have observed similar activations.

\begin{figure}[!htp]
\vskip 0.2in
 \centering
    \subfigure{
    \includegraphics[width=0.8in]{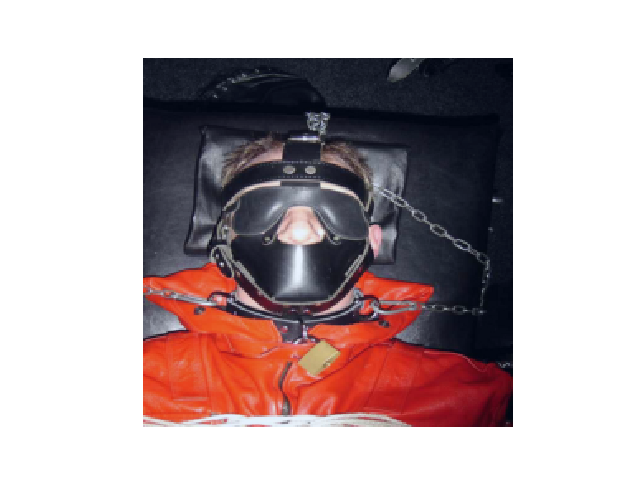}
    \includegraphics[width=0.8in]{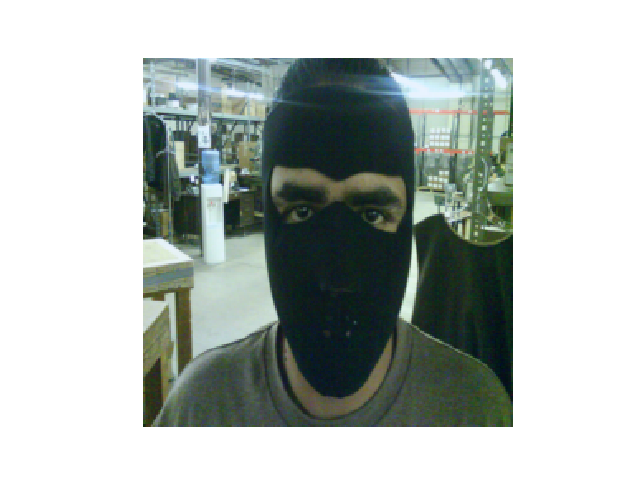}
    \includegraphics[width=0.8in]{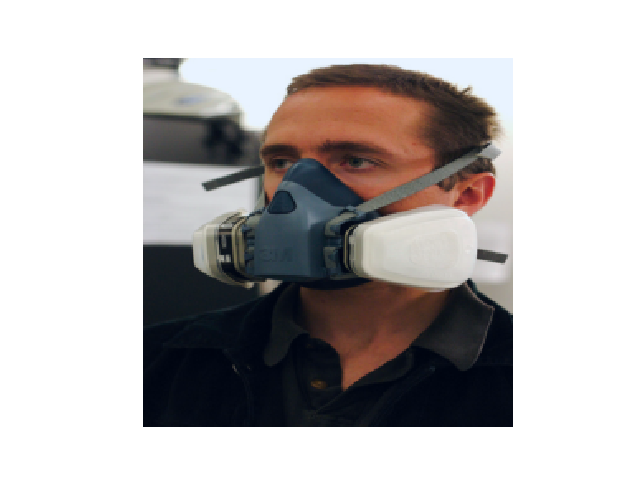}
    \includegraphics[width=0.8in]{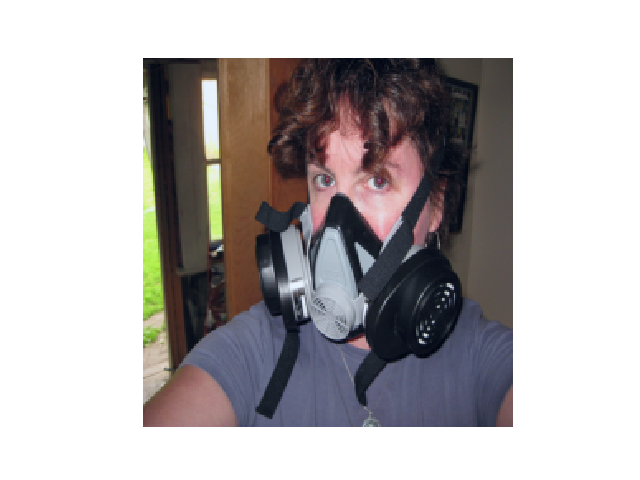}
    }\\
    \subfigure{
    \includegraphics[width=0.8in]{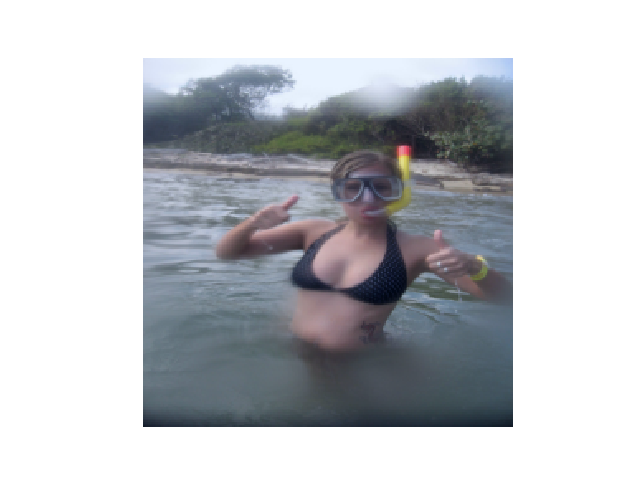}
    \includegraphics[width=0.8in]{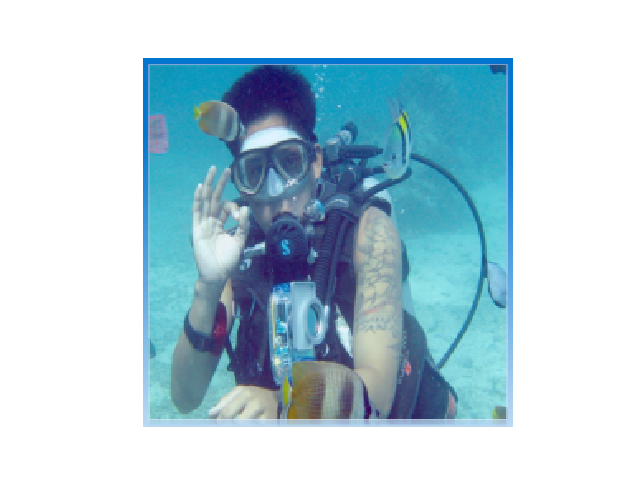}
    \includegraphics[width=0.8in]{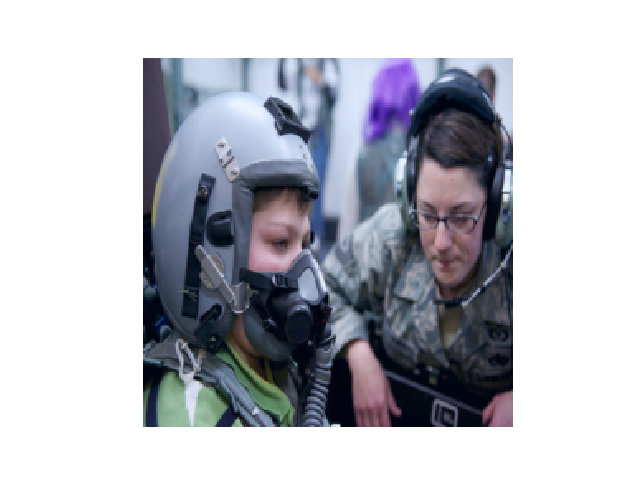}
    \includegraphics[width=0.8in]{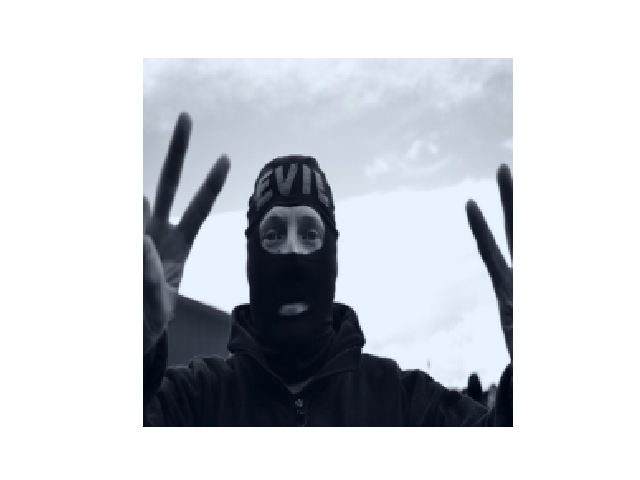}
    }
    \subfigure{
    \includegraphics[width=\columnwidth]{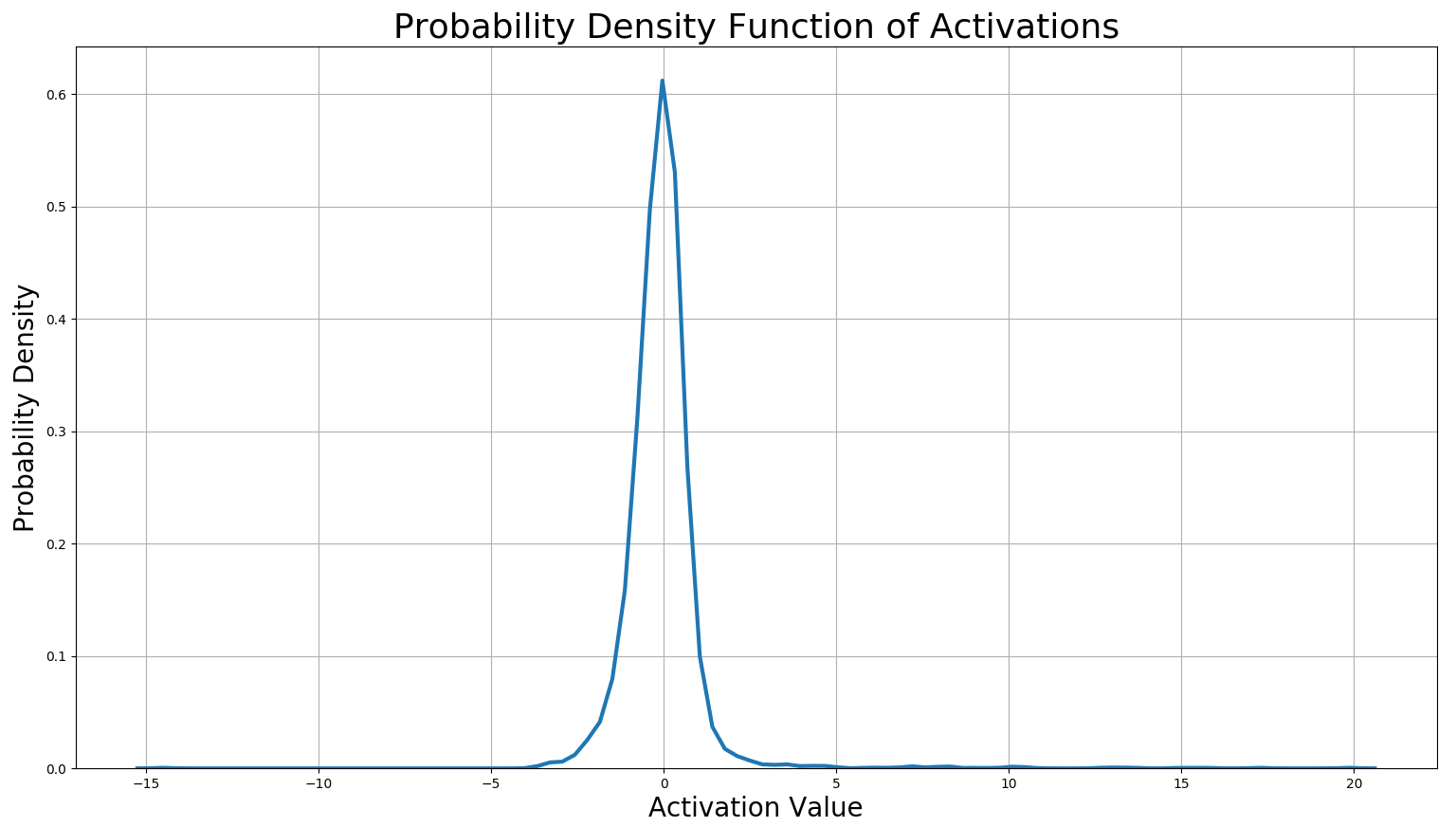}
    }
    \caption{The bottom subfigure displays the density plot of the activations in the 453th kernel of the 13th (last convolutional) layer of VGG-16-BN. As one can see, activations follow Gaussian distribution apart from the extreme outliers, meanwhile the top two rows contain images which resulted activations above $14\sigma$.}
\label{FigVGG14dist}
\end{figure}

To demonstrate that apart from these outliers the activations are indeed Gaussian we have investigated the distribution of a randomly selected (453th) kernel of the 13th layer (last convolutional layer) of VGG-16-BN. We have plotted the distribution of the activations of this single convolution kernel after normalization on the whole validation set which can be seen in Figure \ref{FigVGG14dist}. This distribution can be considered Gaussian with rare outliers, but these outliers can have extremely large values.
To demonstrate the specificity of this randomly chosen kernel, we have also selected all samples from the validation set of ImageNet which results an activation larger than $14\sigma$. Altogether there were 51 such images and all of them contained people with masks. Some randomly selected samples from these 51 images are also displayed on Figure \ref{FigVGG14dist}.
51 images out of the 50,000 samples of the validation set  means that in case of batches of 16, the probability that such image will be present in the mini-batch is 0.01632.
Out of every 61 mini-batch there will be one, where extremely large activations (above $14\sigma$) will be present, therefore  distorted mean and variance values will be calculated for the activations forming the Gaussian like activations.

\begin{figure}[ht]
\vskip 0.2in
\begin{center}
\centerline{\includegraphics[width=\columnwidth]{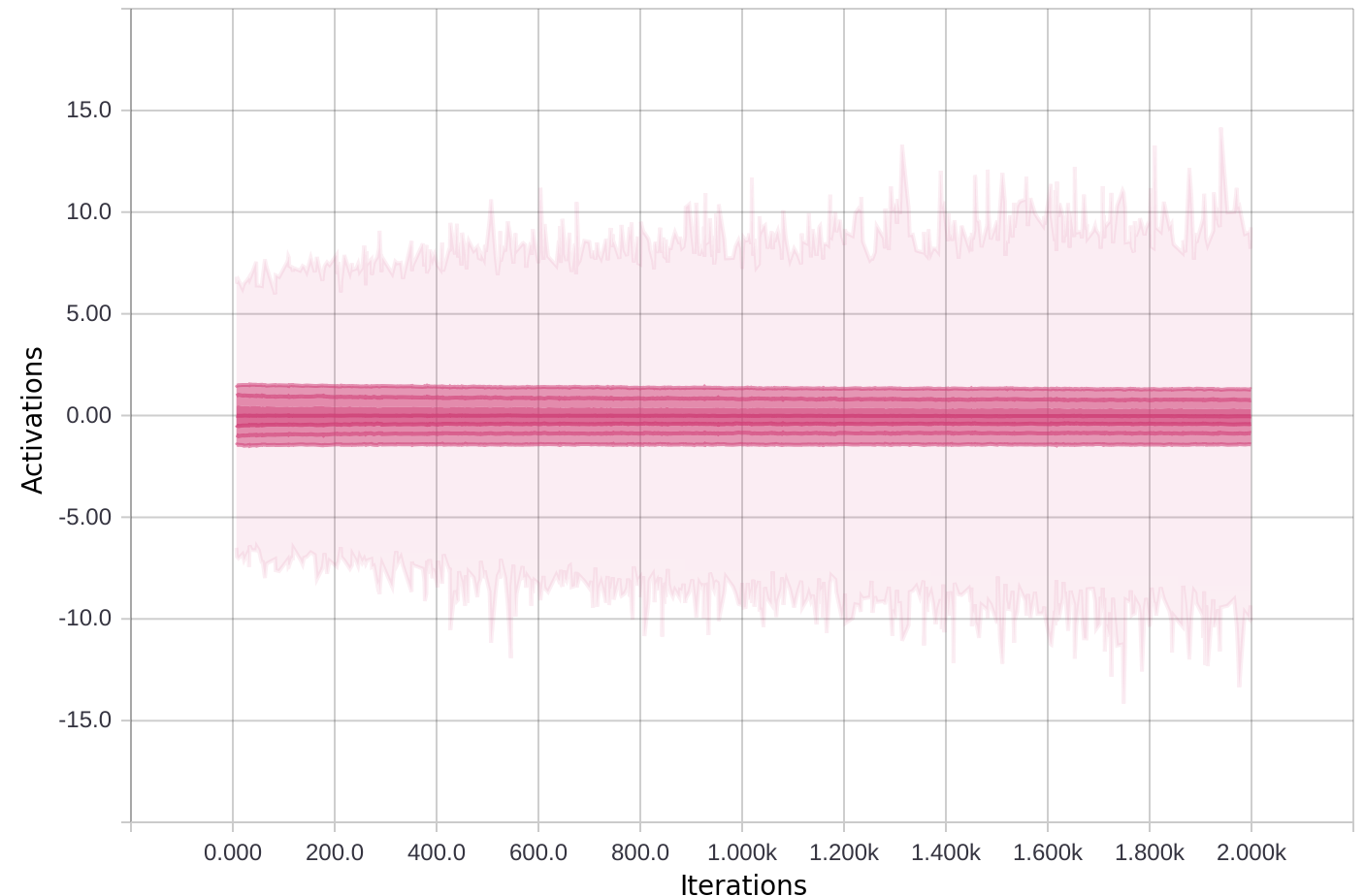}}
\caption{This figure depicts the distribution of activations of the third convolutional layer in AlexNet on the CIFAR-10 dataset after batch-normalization during training in the first 2000 iterations (with batches of 64).  The X axes contains the training iterations; the Y axes contains the value of the activations. The bold lines represent $25\%, 50\%, 75\%$ of the values, and the fourth lines represent the minimum and maximum values. As one can see activations above $10\sigma$ appear in this layer.}
\label{FigureCifarActivations}
\end{center}
\vskip -0.2in
\end{figure}

We have also investigated network activations during training in the AlexNet architecture on the CIFAR-10 dataset. The activations after batch normalization (again without applying the $\gamma$ and $\beta$ parameters) in the third convolutional layer are displayed on Figure \ref{FigureCifarActivations}. As we can easily see, larger and larger activations appear in the network during training.

From the previously presented examples, we can easily see that if we disregard these outliers, activations indeed form a Gaussian distribution, but including these samples can heavily change the calculated mean and variance values during normalization.
This problem is further exacerbated by the fact that training in case of complex networks and datasets happens in mini-batches.
In most mini-batches, activations will have Gaussian distribution, but once a specific extreme activation appears in a channel, it can drastically alter the expected value and the variance which are used for normalization.

To overcome this problem we will introduce filtered batch normalization which removes these outliers before the mean and variance calculation, therefore resulting more consistent distributions over training.

\section{Filtered Batch Normalization}\label{SecFilteredBatchNorm}

At the creation of filtered batch normalization our aim was to design an algorithm that would filter out outliers from a distribution which can appear with low probability in mini-batches, but do not modify the mean and variance values if the input is a perfect Gaussian distribution without outliers.

We were investigating commonly applied methods for robust mean and variance calculation \cite{yuan1998robust}, throwing out the highest and lowest $k$-percent samples from the data or applying winsorization \cite{kokic1994optimal}, substituting these samples with other values. Unfortunately, getting rid of samples or replacing them will change the variance in those cases where the activations do not contain outliers, determining the optimal value of $k$ is difficult and the additional sorting operation of the samples requires  $\mathcal{O}(n\log{}n)$ operations.

In the first step of the algorithm we calculate $\mu_i$ and $\sigma_i$ values similarly as in equation \ref{EQMean} and \ref{EQVar}, but we do not use these values directly for normalization. We create a Gaussian candidate distribution $\hat{x'}_i$ which might contain outliers, but has zero mean and variance of one:

\begin{equation}
    \hat{x'}_i =  \frac{1}{\sigma_i} (x_i - \mu_i) 
\end{equation}

Based on this Gaussian candidate, we create a mask ($f(x_k)$) to select those values which are only less than $T_{\sigma}$ distance from the mean value:

\begin{equation}
f(x_k)=\begin{cases}
1 & \text{ if } -T_{\sigma}  \leq \hat{x'}_k \leq T_{\sigma}\\ 
0 & \text{ if } \hat{x'}_k< -T_{\sigma} \vee T_{\sigma}< \hat{x'}_k
\end{cases}
\end{equation}

$T_{\sigma}$ is a hyperparameter of the algorithm and the performance of the algorithm does not depend heavily on its value. As it can be seen from the previous example, $T_{\sigma}=7$ can filter out most outliers in the data.

We use this mask to calculate the mean and the variance only including those which are not considered outliers (which are inside the $\pm T_{\sigma}$ band of the mean value).

\begin{equation}
    \mu'_i = \frac{1}{ {\sum\limits_{k \in S_i}} f(x_k)}  \sum_{k \in S_i} f(x_k)x_k
\end{equation}

\begin{equation}
\sigma'_i = \sqrt{\frac{1}{\sum\limits_{k \in S_i}{f(x_k)}}  \sum_{k \in S_i} f(x_k) (x_k - \mu'_i)^2 + \epsilon } 
\end{equation}


$\mu'_i$  and $\sigma'_i$ values are used to transform the activations which can be calculated similarly to equation \ref{EqBatchNorm}:

\begin{equation}
    y'_i = \gamma \frac{(x_i - \mu'_i)}{\sigma'_i}  +  \beta
\end{equation}

If our loss at the end of the filtered batch normalization layer is defined as $\ell$, backpropagation of the filtered batch normalization layer can be defined by the following equations:

\begin{equation}
\resizebox{.9\hsize}{!}{$
\frac{\partial \ell}{ \partial {\sigma'}{_i}{^2} } =
\sum\limits_{k \in S_i} f(x_k)
\frac{\partial \ell}{\partial y'_k }\gamma
(x_k-\mu'_i)
\frac{-1}{ 2 } 
( {\sigma'}{_i}{^2} + \epsilon)^{\frac{-3}{2 }}
$}\end{equation}

\begin{equation}
\resizebox{.9\hsize}{!}{$
\frac{\partial \ell}{ \partial \mu'_i } =
\sum\limits_{k \in S_i}{ f(x_k)}
\frac{\partial \ell}{\partial y'_k }\gamma
\frac{-1}{ \sqrt{ {\sigma'}{_i}{^2} + \epsilon} } 
+\frac{\partial \ell}{ \partial {\sigma'}{_i}{^2} }
\frac{\sum\limits_{k \in S_i} -2f(x_k)(x_k - \mu'_i)}{ \sum\limits_{k \in S_i} f(x_k)}
$}\end{equation}

\begin{equation}
\resizebox{.9\hsize}{!}{$
\frac{\partial \ell}{\partial x_i } = \frac{\partial \ell}{\partial y'_i }\gamma \frac{1}{\sqrt{{\sigma'}{_i}{^2}+\epsilon}} + \frac{\partial \ell}{\partial {\sigma'}{_i}{^2} } 
 \frac{2 f(x_i)(x_i-\mu'_i)}{ \sum\limits_{k \in S_i}{f(x_k)} } +  \frac{\partial\ell}{\partial\mu'_i}  \frac{1}{ \sum\limits_{k \in S_i}{f(x_k)} } 
$}\end{equation}

\begin{equation}
\frac{\partial \ell}{\partial \beta } =  \sum\limits_{k \in S_i}{ f(x_k)\frac{\partial \ell}{\partial y'_k } }
\end{equation}
\begin{equation}
\frac{\partial \ell}{\partial \gamma } =  \sum\limits_{k \in S_i}{ f(x_k)\frac{\partial \ell}{\partial y'_k }   \frac{(x_k - \mu'_k)}{\sigma'_k}  }
\end{equation}

Computation wise, the calculation of this method requires only an additional mean and variance calculation in training, which is less expensive ($\mathcal{O}(2n)$) than sorting the samples. Furthermore, the mean and variance, which can be used directly in inference can be learned and smoothed during training with moving averages through iterations as it is done in batch re-normalization. This means that in inference the application of filtered batch normalization does not have any additional computational overhead. 

In training, only a minor time increase could be observed: one iteration of training with VGG-16 without batch normalization took in average $148 ms$ on an NVIDIA GTX 2080 TI using Pytroch, which increased to $164 ms$ with batch normalization and finally resulted $172 ms$ using filtered batch normalization.

The additional hyperparameter of the algorithm $T_{\sigma}$ can be tuned fairly easily. We were typically using values between two and seven and have not observed major changes in accuracy.

We also would like to emphasize that this approach can be used together with other methods as well, the important part is the filtering step and it can be similarly applied in filter response normalization \cite{singh2019filter} or group normalization \cite{wu2018group}.

\section{Results}\label{SecResults}
Here we will introduce our main findings on commonly applied network architectures and datasets. We will only list the most important hyperparameters of our training algorithms, but we would like to emphasize that all of our training scripts and codes !!!are shared as supplementary material and will be shared publicly in the final version.
Also in all comparisons only the normalization method was changed, all other hyperparameters from batch size to optimization algorithms remained the same.
In all our experiments we were using batch re-normalization: smoothing was used during training (with an $\alpha$ value of $0.1$) to determine the means and variances of the batches and these values were calculated during training and used as constant values for normalization during inference and evaluation.

\subsection{MNIST}

We have selected the LeNet-5 architecture and the MNIST dataset as a proof of concept to validate our method. This simple dataset allows for detailed investigations of our method using various parameters.

We have investigated the effect of $T_{\sigma}$ parameter on the performance of our algorithm and compared it to the traditional approach without using batch normalization which we will refer to as NO-BN and also to the built-in batch normalization algorithm of Pytorch (it uses batch re-normalization and stores the momentum of the mean and variance values) which we will refer to as (BN).
We have used a single batch normalization layer between the last two fully connected layers to demonstrate its effect without major modifications in the network.

\begin{figure}[!htp]
\vskip 0.2in
\begin{center}
\centerline{\includegraphics[width=\columnwidth]{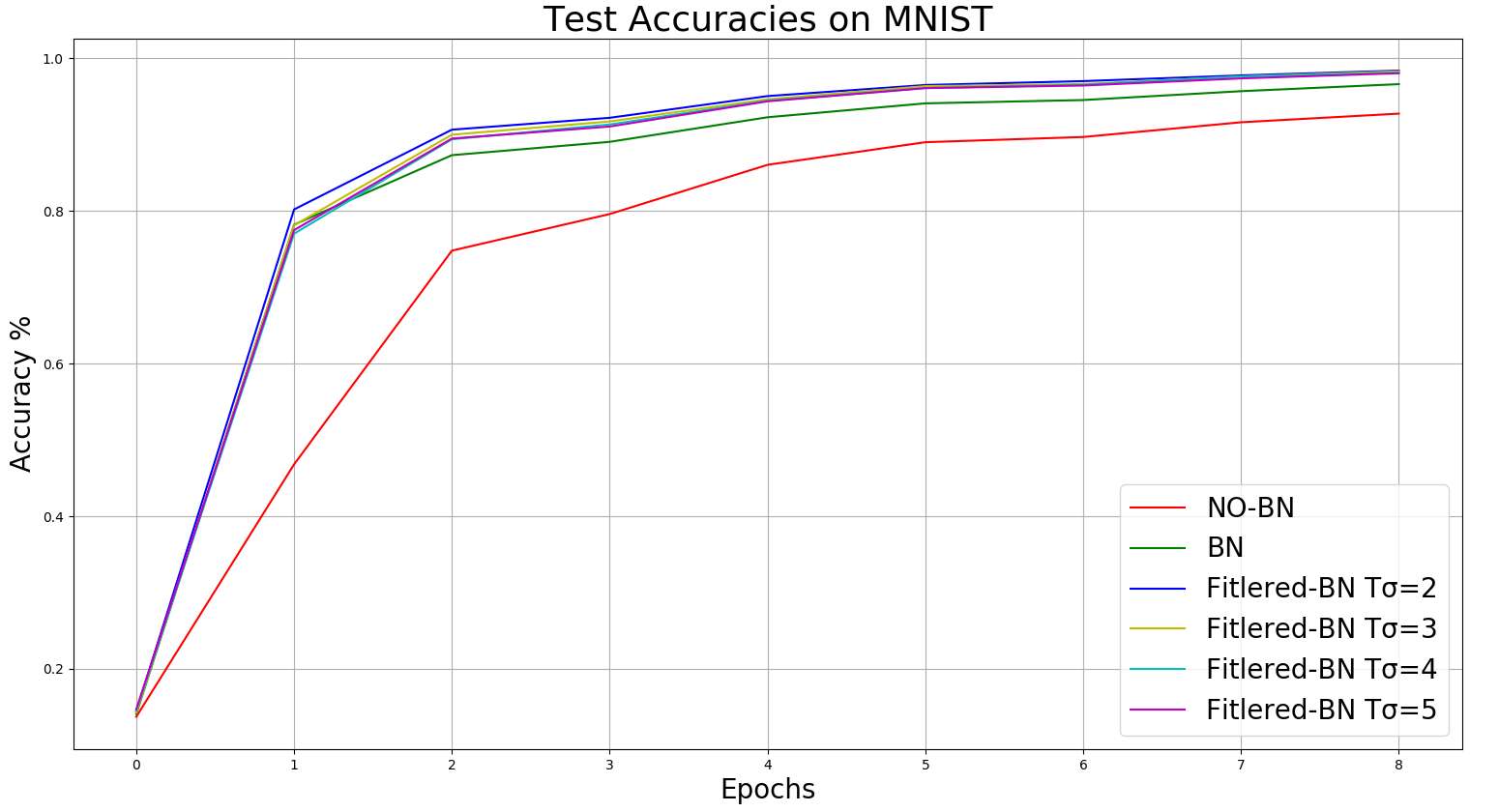}}
\caption{This figure depicts the test accuracies of our algorithm (Filtered-BN) with various $T_{\sigma}$ parameters compared to the Pytorch built-in batch normalization (BN, green) and no batch normalization (NO-BN, red). Each of the displayed results is the average of ten independent trainings, which were executed with batches of 256.}
\label{FIGMNISTResults}
\end{center}
\vskip -0.2in
\end{figure}

The results can be seen in Figure \ref{FIGMNISTResults}, where filtered batch normalization results higher test accuracies and faster convergence. Also one can notice that changing $T_{\sigma}$ parameter would only affect the test results slightly. Of course with a large enough sigma which contains all outliers (e.g. $T_{\sigma}=100$) we would get the original BN method back.

\begin{figure}[!htp]
\vskip 0.2in
\begin{center}
\centerline{\includegraphics[width=\columnwidth]{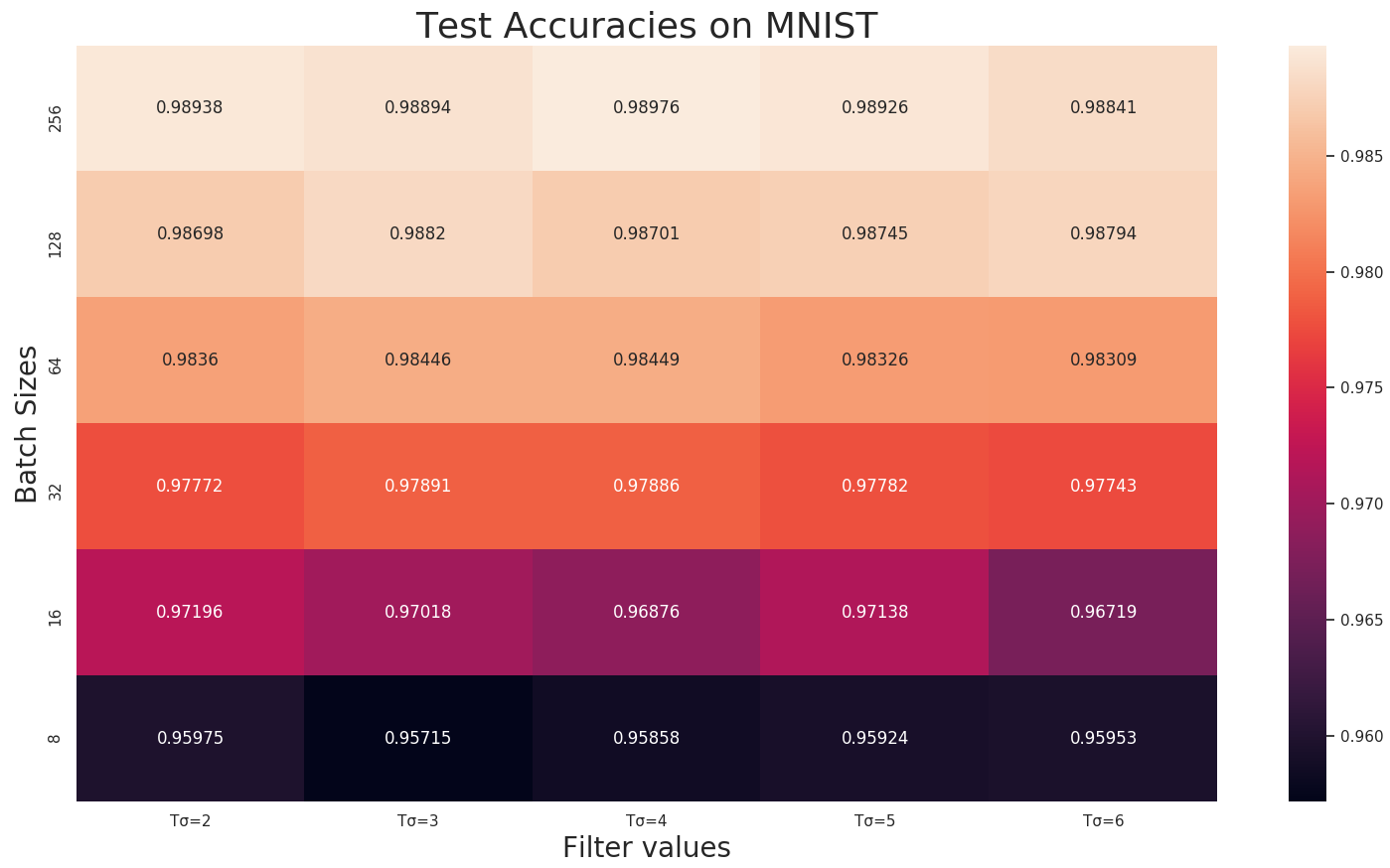}}
\caption{Test accuracies on the MNIST test set with different mini-batch sizes (rows) and $T_{\sigma}$ values (columns) after 1000 iterations. As it can be seen accuracy values are fairly robust against the changes of $T_{\sigma}$, but depend heavily on mini-batch sizes. }
\label{FigMnistBatches}
\end{center}
\vskip -0.2in
\end{figure}

We have also investigated how the results depend on batch sizes and the hyperparameter $T_{\sigma}$. The results are depicted in Figure \ref{FigMnistBatches}. As it can be seen the accuracy after a given number of training iterations depends heavily on the size of the mini-batch, but changing the parameter $T_{\sigma}$ does not cause drastic alterations in accuracy.

\begin{figure}[!htp]
\vskip 0.2in
 \centering
    \subfigure{
    \includegraphics[width=\columnwidth]{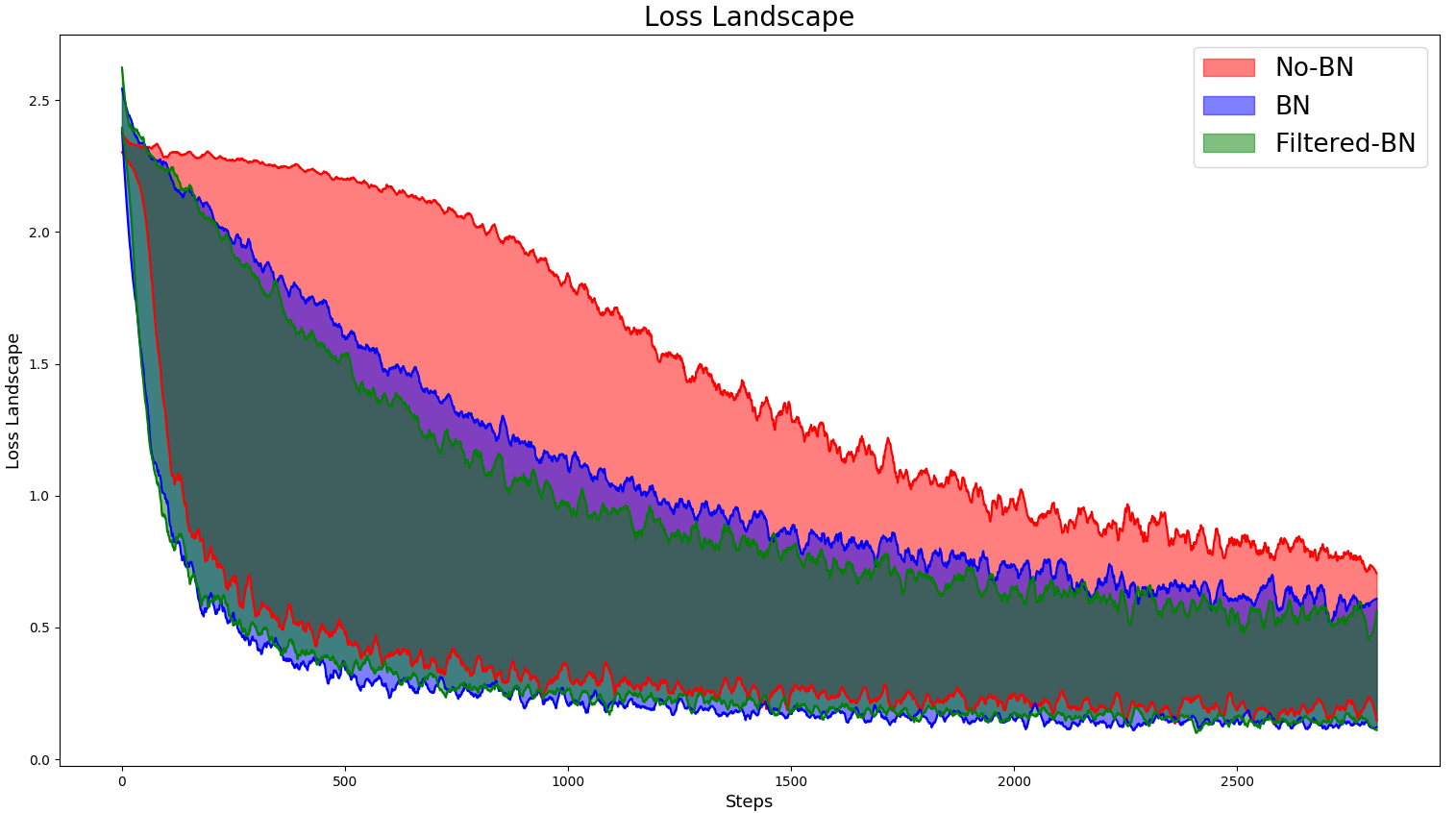}}
    \\ \subfigure{
    \includegraphics[width=\columnwidth]{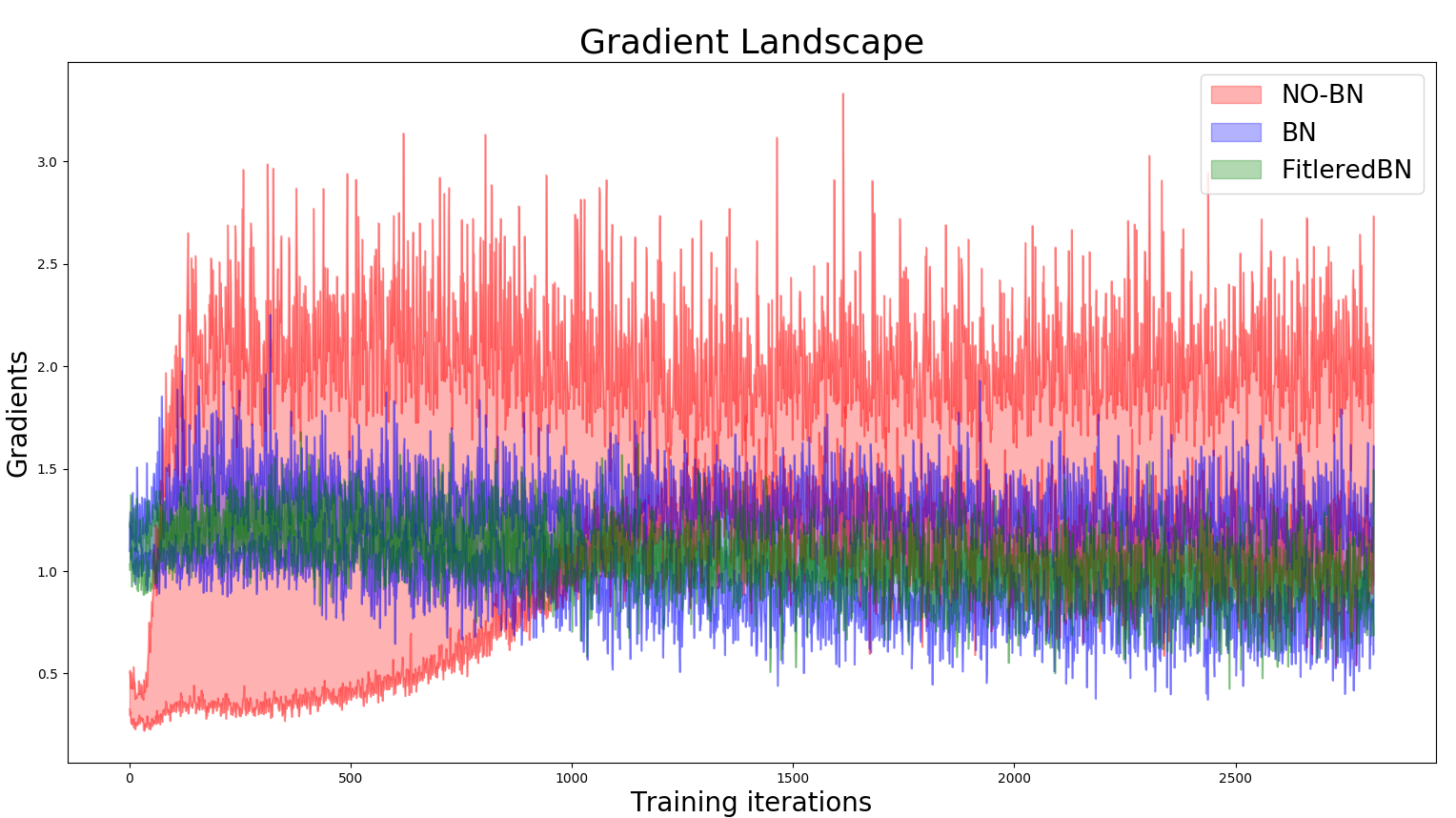}
    }
    \caption{The loss and gradient landscapes for the first 3000 iterations on the MNIST dataset with batches of 64 are depicted on these figures. The left figure (Loss Landscape) displays the variance in loss as we move in the gradient direction at a particular training step. Meanwhile the right figure (Gradient Landscape) depicts the variance in $\ell_2$ changes of the gradients in a similar setup. The variances were calculated between the current gradient and loss and as if a step would be made toward the gradient with step-size 0.02, 0.01, 0.005 and 0.001.
    As it can be seen the variance in the loss and gradient values are smaller using our method resulting smoother loss and gradient surfaces.}
\label{FigMnistLossLandscape}
\vskip -0.2in
\end{figure}

In \cite{santurkar2018does} a thorough investigation has been conducted searching for the reasons behind the success of BN.  The authors demonstrate that inner covariate shift has only a tenuous effect and BN mostly helps by creating a smooth loss and gradient landscape. To investigate this, we have also compared the loss and gradient surface of our method similarly as the authors did in \cite{santurkar2018does}. The results can be seen on Figure \ref{FigMnistLossLandscape}. As we can observe, even in the case of a simple network and dataset, filtered-BN (in this experiments with $T_{\sigma}=2$) results a smoother loss and gradient landscape. 

\subsection{CIFAR-10}

We have also investigated the AlexNet architecture on CIFAR-10.
We have examined three different architectures: the vanilla implementation of AlexNet without any normalization, one containing the built-in batch normalization after every layer (except the logit)  and one containing filtered batch normalization (with $T_{\sigma}=2$). All the training parameters, optimizer (gradient descent with momentum) and batch size (128) were the same.

\begin{figure}[!htp]
\vskip 0.2in
\begin{center}
\centerline{\includegraphics[width=\columnwidth]{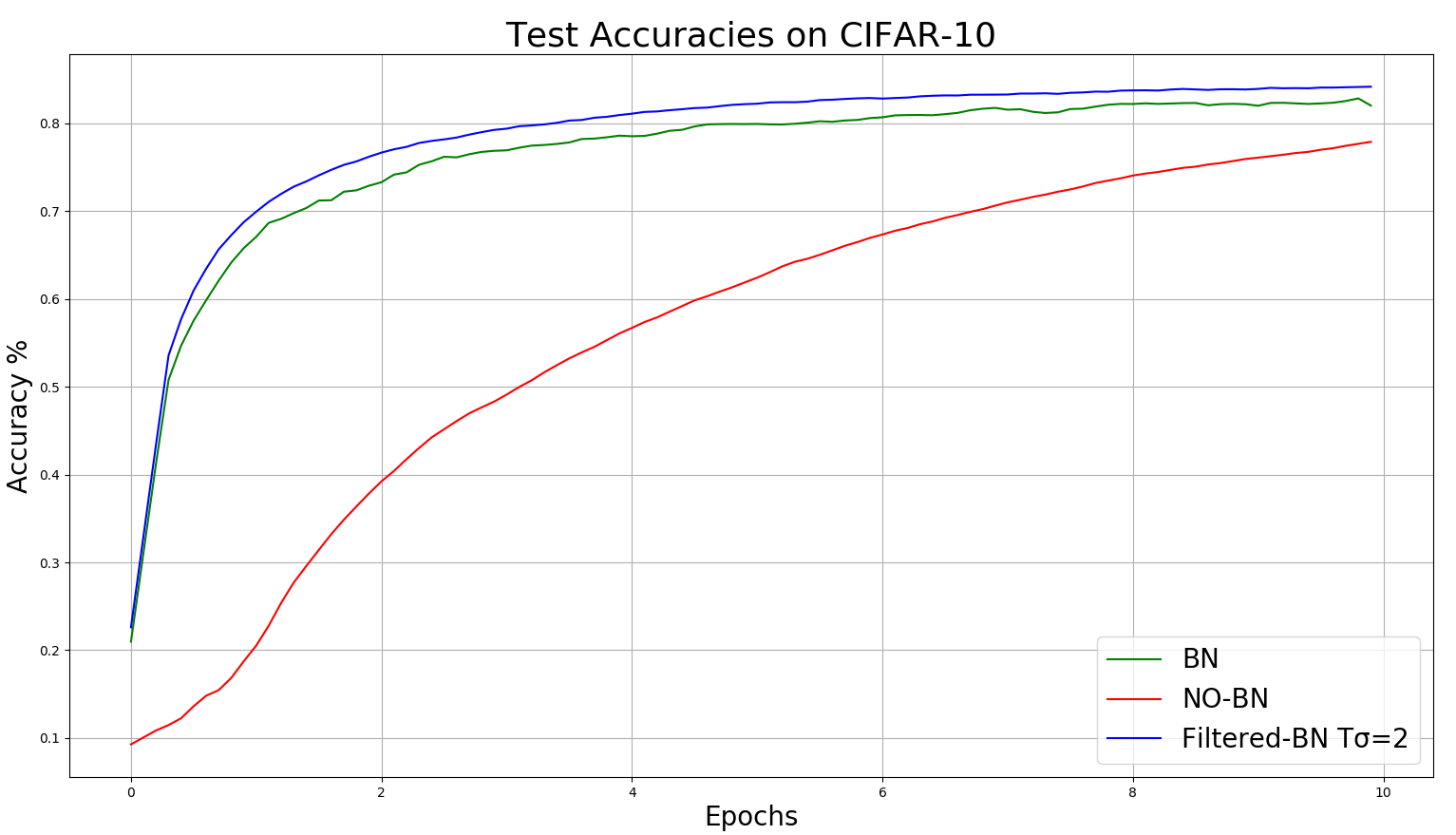}}
\caption{This Figure demonstrated test accuracies on CIFAR-10 with the AlexNet architecture using three different normalization methods: no normalization (NO-BN, red), traditional BN (green) and filtered BN with parameter $T_{\sigma}=2$ (blue).}
\label{FigCifarAcc}
\end{center}
\vskip -0.2in
\end{figure}

The original $32\times32$ images of CIFAR-10 were rescaled to $227\times227$ to ensure the appropriate input dimensions for AlexNet.
Ten independent runs were executed and the averaged test accuracies can be seen in Figure \ref{FigCifarAcc}.
The network without BN achieved a test accuracy of $77\%$ in average over 10 independent runs, with the built-in implementation of BN it achieved $82\%$ and with filtered batch normalization the network has reached $84\%$ accuracy.

Comparing the mean and variance values is difficult since they change between iterations because of two reasons. The difference can be caused by inner covariate shift and also by the variance of the input features in each mini-batch. Our aim was to minimize the later effect to ensure consistent moments for normalization which are close to the global mean and variance. Unfortunately, these values can not be calculated on the entire dataset during training.

To investigate the consistency of the mean and variance values of our approach which would imply a stable distribution, we have trained Alexnet using regular batch normalization with large batches (128) on CIFAR-10 (we will refer to this as BN128). In each training step, we have randomly selected a smaller mini-batch (16 samples) out of these 128 instances, then we have calculated the moments on this small mini-batch on the same network using batch normalization (we will refer to this as BN16) and filtered batch normalization (FBN16).

\begin{figure}[!htp]
\vskip 0.2in
 \centering
    \subfigure{
    \includegraphics[width=\columnwidth]{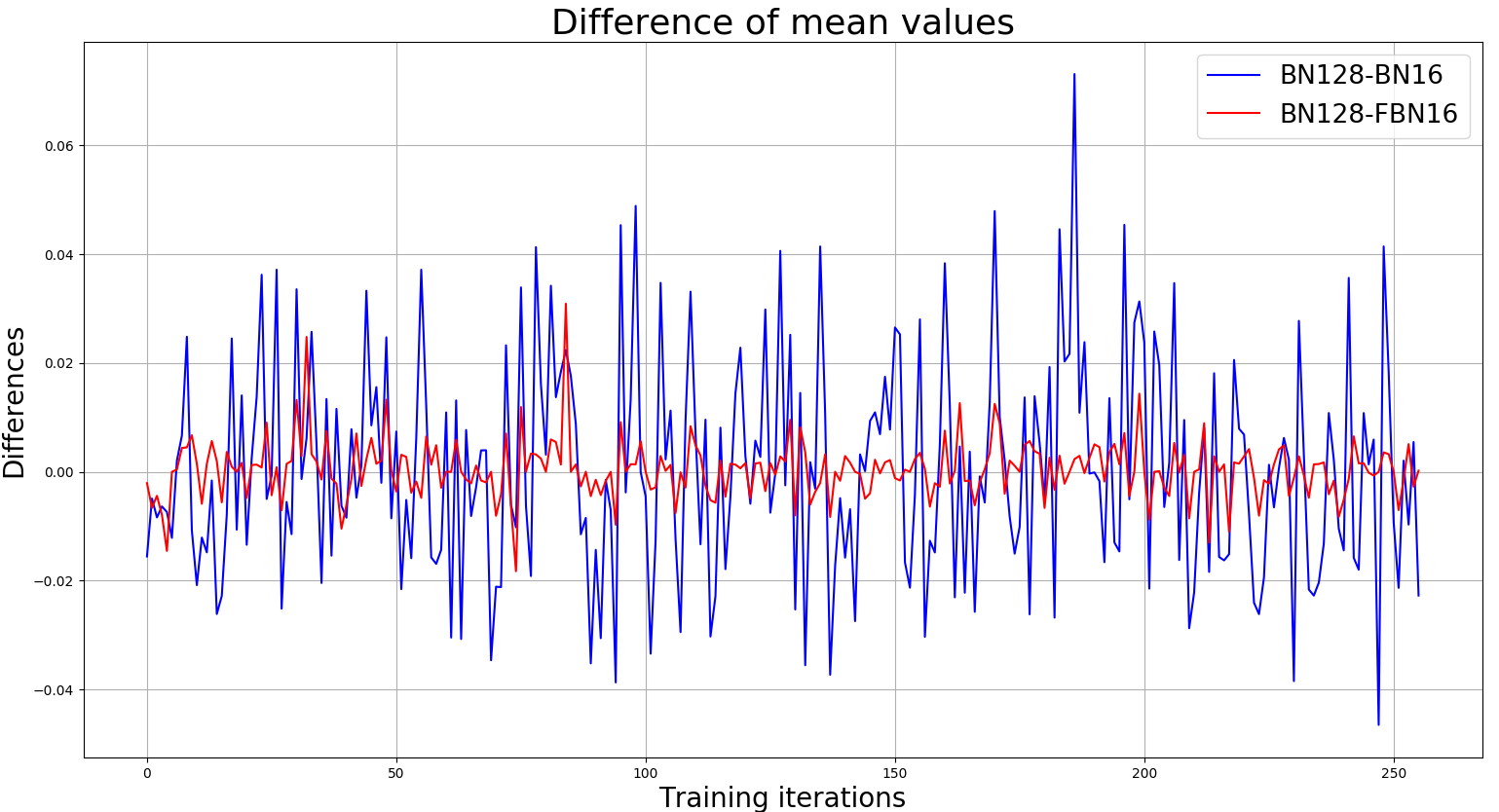} }\\
    \subfigure{\includegraphics[width=\columnwidth]{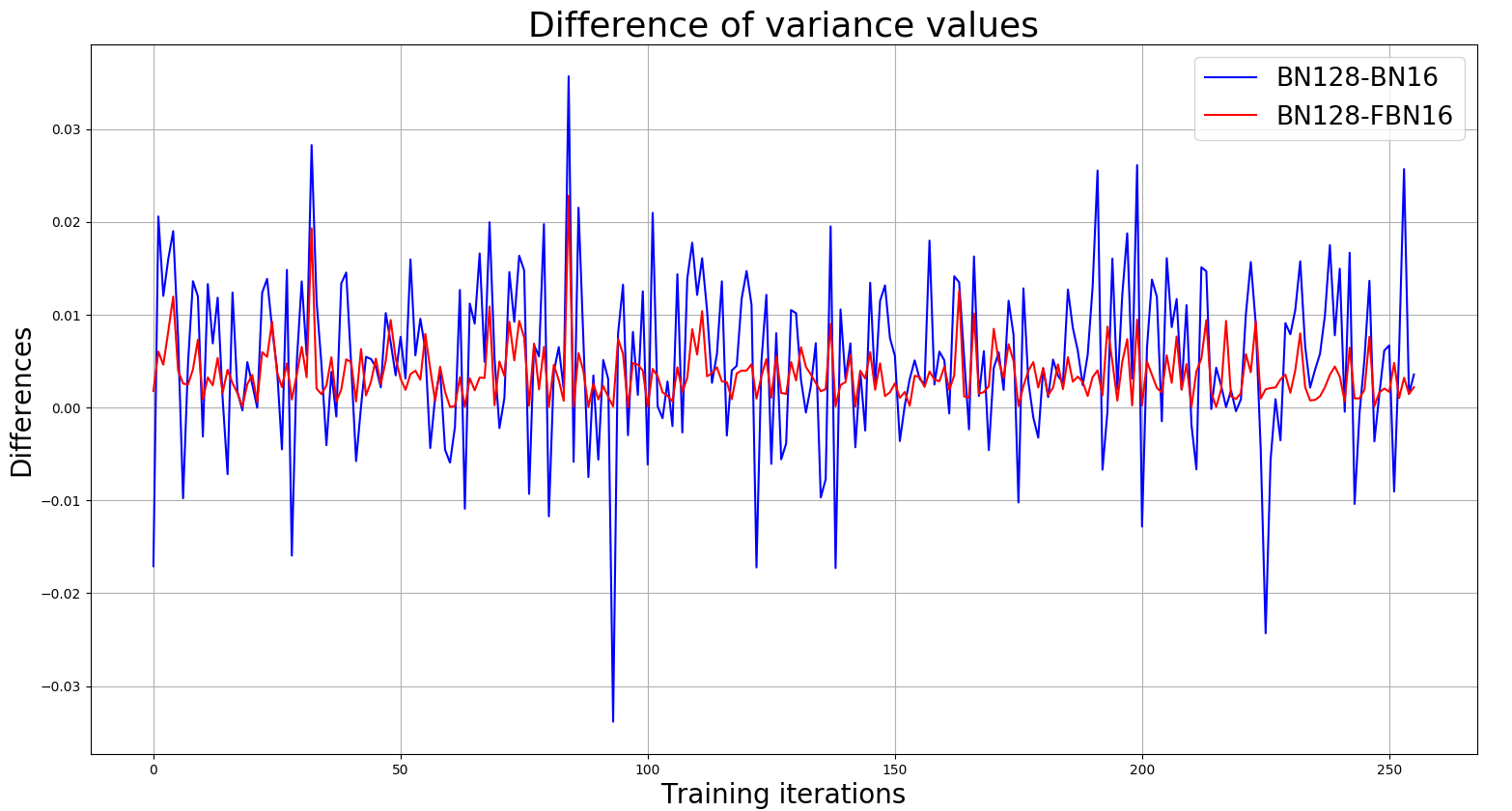}
    }
    \caption{This figure depicts the mean and variance differences between traditional BN applied with mini-batches of 128 and mini-batches of 16 (blue), BN with batches of 128 and filtered BN with mini batches of 16 (red). As it can be clearly seen filtered batch normalization approximates the mean and variance values of larger batch sizes better than traditional BN.}
\label{FigCIFARSmallLargeBatch}
\vskip -0.2in
\end{figure}

We have considered a larger mini-batch of 128 samples as a reference for the consistent mean and variance values\footnote{Although divergence can happen even with large batch sizes but one can assume that larger values approximate the mean and variance values of the whole dataset better.} and compared them to the mean and variance values of the smaller mini-batch calculated by either regular BN or filtered BN. The difference between the moments of the large and the small batches using regular  and filtered BN can be seen in Figure \ref{FigCIFARSmallLargeBatch}.

\subsection{ImageNet}
\begin{figure}[!htp]
\vskip 0.2in
\begin{center}
\centerline{\includegraphics[width=\columnwidth]{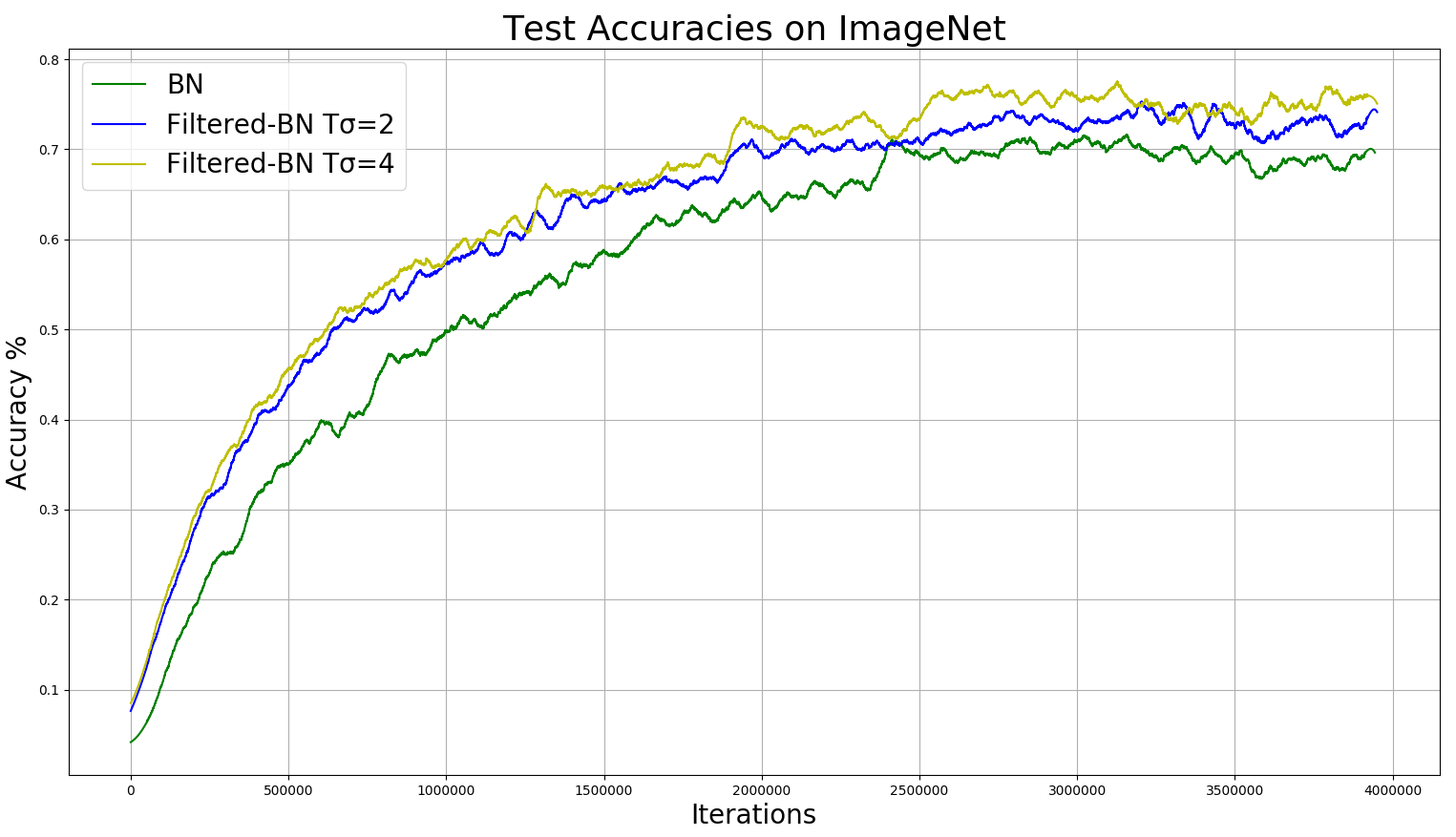}}
\caption{This figure depicts the accuracies on the validation set of ImageNet using the VGG-16-bn architecture with the traditional BN and filtered BN with $T_{\sigma}$ equals two and four. As it can be seen from the results filtering out outliers in the activations has increased the convergence speed and overall accuracy of the network.}
\label{FigimageNet}
\end{center}
\vskip -0.2in
\end{figure}

\begin{figure}[!htp]
\vskip 0.2in
\begin{center}
\centerline{\includegraphics[width=\columnwidth]{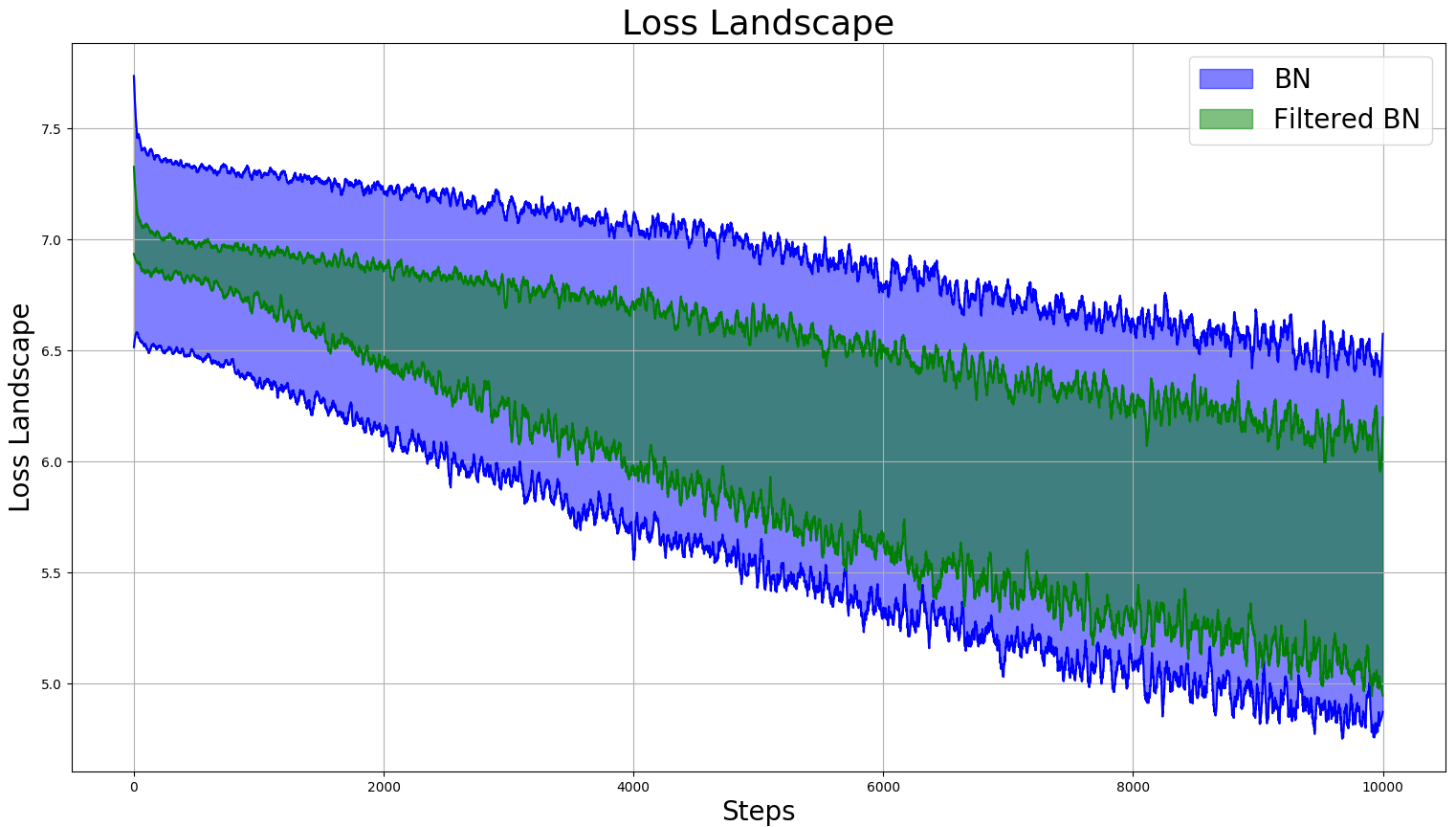}}
\caption{We have investigated how the loss function of the network varies in training according to training steps with different sizes (0.02, 0.01, 0.005, 0.001). These variances selecting the minimal and maximal values are depicted on this figure for regular batch normalization and filtered batch normalization with $T_{\sigma}=2$ for the first 10k iterations of training with batches of 16.}
\label{FigimageNetLoss}
\end{center}
\vskip -0.2in
\end{figure}

We have also investigated the effect of filtered BN on the VGG-16 architecture on the ImageNet 2012 dataset. Normalizing layers were added after every convolution and fully connected layer (except the logit layer). Training was implemented with batches of 16 and the top-1 accuracies at different iterations on the validation set can be seen on Figure \ref{FigimageNet}. In four million iterations the network has reached the following top-1 accuracies on the validation set: $69\%$ with the application regular BN and $73\%$ and $74\%$ with the application of filtered batch normalization, with the corresponding $T_{\sigma}=2$ and $T_{\sigma}=4$ parameters. Meanwhile the loss landscape of VGG-16 on ImageNet with the two different normalization methods can be seen on Figure \ref{FigimageNetLoss}.

\subsection{Group Normalization}

\begin{figure}[!htp]
\vskip 0.2in
\begin{center}
\centerline{\includegraphics[width=\columnwidth]{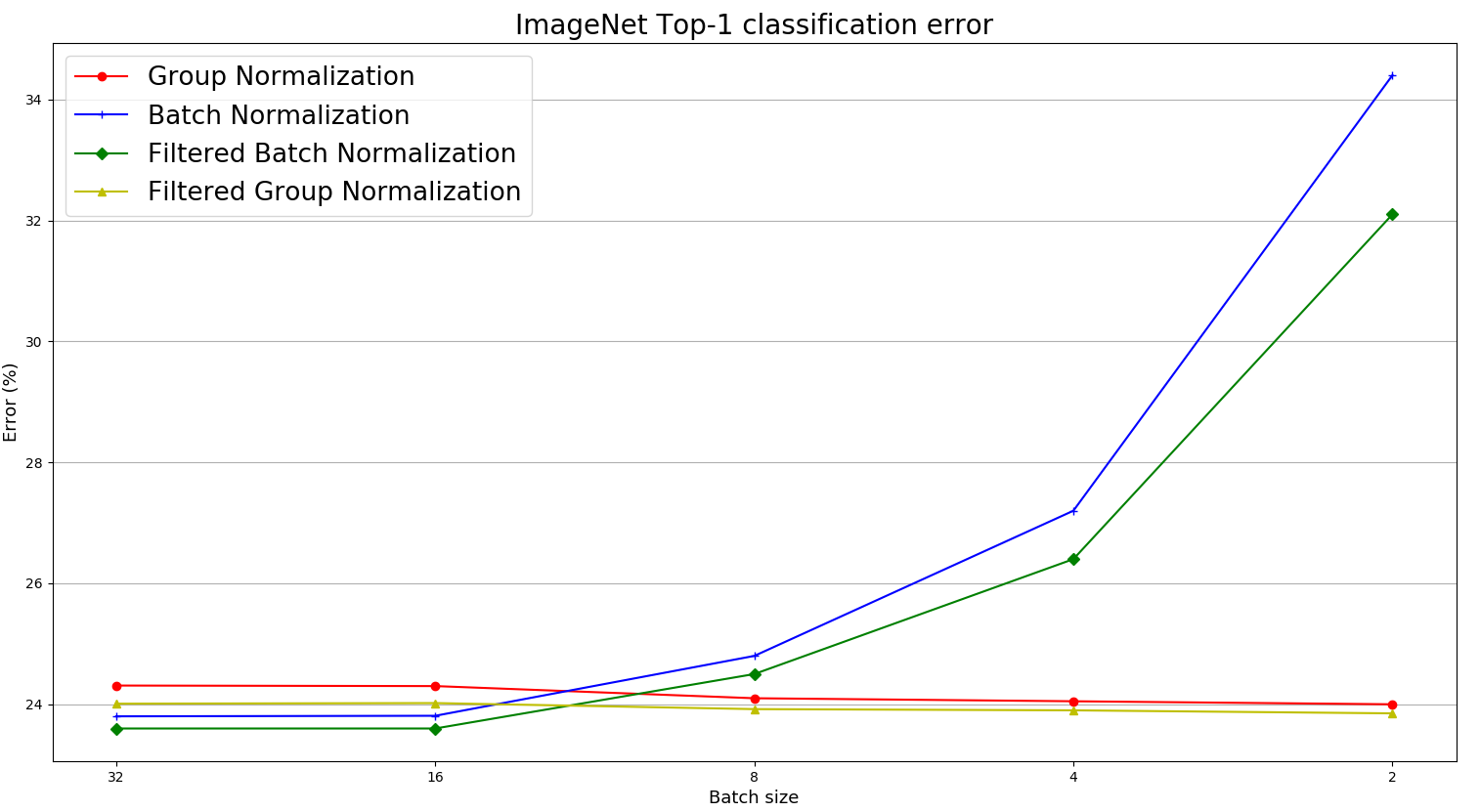}}
\caption{Top-1 error rate on the validation set of ImageNet using ResNet-50 and four different normalization method. As it can be seen the filtered approaches outperformed their original counterparts with all batch sizes.}
\label{FigGrouNorm}
\end{center}
\vskip -0.2in
\end{figure}

It was recently demonstrated that group normalization has superior performance in case of small mini-batches over BN.
Group normalization uses the same method for mean and variance calculation. The only difference is the selection and grouping of the elements for normalization ($S_i$). In this method elements are selected across channels and positions, but from a single instance. Thus, our filtering method can be applied with group normalization after element selection. Equation 1 and 2 of the original paper \cite{wu2018group} can easily be changed according to the method described in Section \ref{SecFilteredBatchNorm}. We will refer to the modified algorithm as filtered group normalization.
To compare these methods we have selected the ResNet-50 architecture and applied batch normalization, filtered batch normalization, group normalization and filtered group normalization on ImageNet with various batch sizes. The top-1 error rates on the validation set are depicted on figure \ref{FigGrouNorm}. As it can be seen from the results, filtered batch normalization resulted the lowest error rate, also in case of small batches, filtered group normalization has outperformed group normalization.

\subsection{Instance segmentation on MS-COCO}

To test our method apart from classification tasks we have also applied it for instance segmentation and object localization on the MS-COCO dataset.
We have selected the Detectron2 \cite{wu2019detectron2} framework for evaluation and used MASK R-CNN with ResNext-101 backbone with feature pyramid network. Training was executed for 270,000 iterations, with 2 images per batch and $T_\sigma$ was set to five. The average precision results are displayed in Table \ref{TableMaskRCNN}. As it can be seen from the results, filtered batch normalization results higher AP at every iteration and also results better final accuracy after 270,000 iterations.

\begin{table}[!t]
\caption{Test accuracies for Mask-RCNN on MS-COCO on Segmentation (Seg) and object detection with bounding boxes (Box) tasks at different iterations (50000, 100000, 150000 and 270000). The columns show mean average precision at $IoU=50:.05:.95$ ($AP$) and average precision at $IoU=0.5$ ($AP_{50}$) with the traditional batch normalization (BN) and filtered batch normalization (F-BN)  }\label{TableMaskRCNN}
\label{sample-table}
\vskip 0.15in
\begin{center}
\begin{small}
\begin{sc}
\begin{tabular}{l|c|c|c|c}
\toprule
 & BN$AP$ & BN$AP_{50}$ & F-BN$AP$ & F-BN$AP_{50}$\\
\midrule
Seg(50k)   & 23.87& 44.56 & 25.41& 45.42\\
Seg(100k)  & 26.86  & 45.55 & 27.34 & 52.40 \\
Seg(150k) & 28.66&  51.80 &  34.15& 55.43\\
Seg(270k)  & 36.47 & 58.07 & 37.06 & 58.92\\
\hline
Box(50k)   & 23.63 & 41.90 & 27.86 & 47.84\\
Box(100k)  & 28.16 & 48.43 &  28.74 & 49.65\\
Box(150k)  & 30.53 & 50.79  &  34.24 & 53.13\\
Box(270k)  & 40.01 & 61.32 & 41.12 & 61.71 \\
\bottomrule
\end{tabular}
\end{sc}
\end{small}
\end{center}
\vskip -0.1in
\end{table}

\section{Conclusion}\label{SecConclusion}
In this paper we have demonstrated that the common assumption, that neural network activations of a selected layer follow Gaussian distribution is not entirely true and contradicts the specificity of neuron and convolutional kernels in deeper layers. We have shown the presence of high activations which can only be explained with unlikely low probabilities using Gaussian distribution.
These, extremely out-of-distribution, seldomly occurring samples can result inconsistent mean and variance values in batch normalization.

We have introduced an algorithm, filtered batch normalization, to filter out these activations resulting faster convergence and higher overall accuracy as we have demonstrated using multiple datasets and network architectures. Our empirical results show that we can create more coherent output distributions in neural network layers by removing these outliers before mean and variance calculation, which results faster convergence and better overall validation accuracies.
We also have to emphasise that comparing to batch normalization, our method adds only a minor computation overhead during training ( $\sim5\%$ in VGG-16), but does not require additional computation in inference mode.

We have also showed that this normalization method results a smoother loss and gradient landscape than batch normalization. We have also demonstrated that our method can be applied with other normalization techniques as well, such as group normalization whose performance could also be increased by filtering out the out-of-distribution activations. 

\section*{Acknowledgements}

This research has been partially supported by the Hungarian Government by the following grant: 2018-1.2.1-NKP-00008 Exploring the Mathematical Foundations of Artificial Intelligence and the support of the grant EFOP-3.6.2-16-2017-00013 is also gratefully acknowledged.

\bibliographystyle{IEEEtran}
\bibliography{batchnorm.bib}

\end{document}


\title{Filtered Batch Normalization\\ -Supplementary material-}

\author{\IEEEauthorblockN{András Horváth, Jalal Al-afandi}
\IEEEauthorblockA{Peter Pazmany Catholic University\\
Faculty Of Information Technology and Bionics\\
Budapest, Hungary\\
Email: horvath.andras@itk.ppke.hu,\\ alafandi.mohammad.jalal@itk.ppke.hu}
}

\maketitle

\section{Network activations}

\begin{figure}[!htp]
\vskip 0.2in
 \centering
    \subfigure{
    \includegraphics[width=\columnwidth]{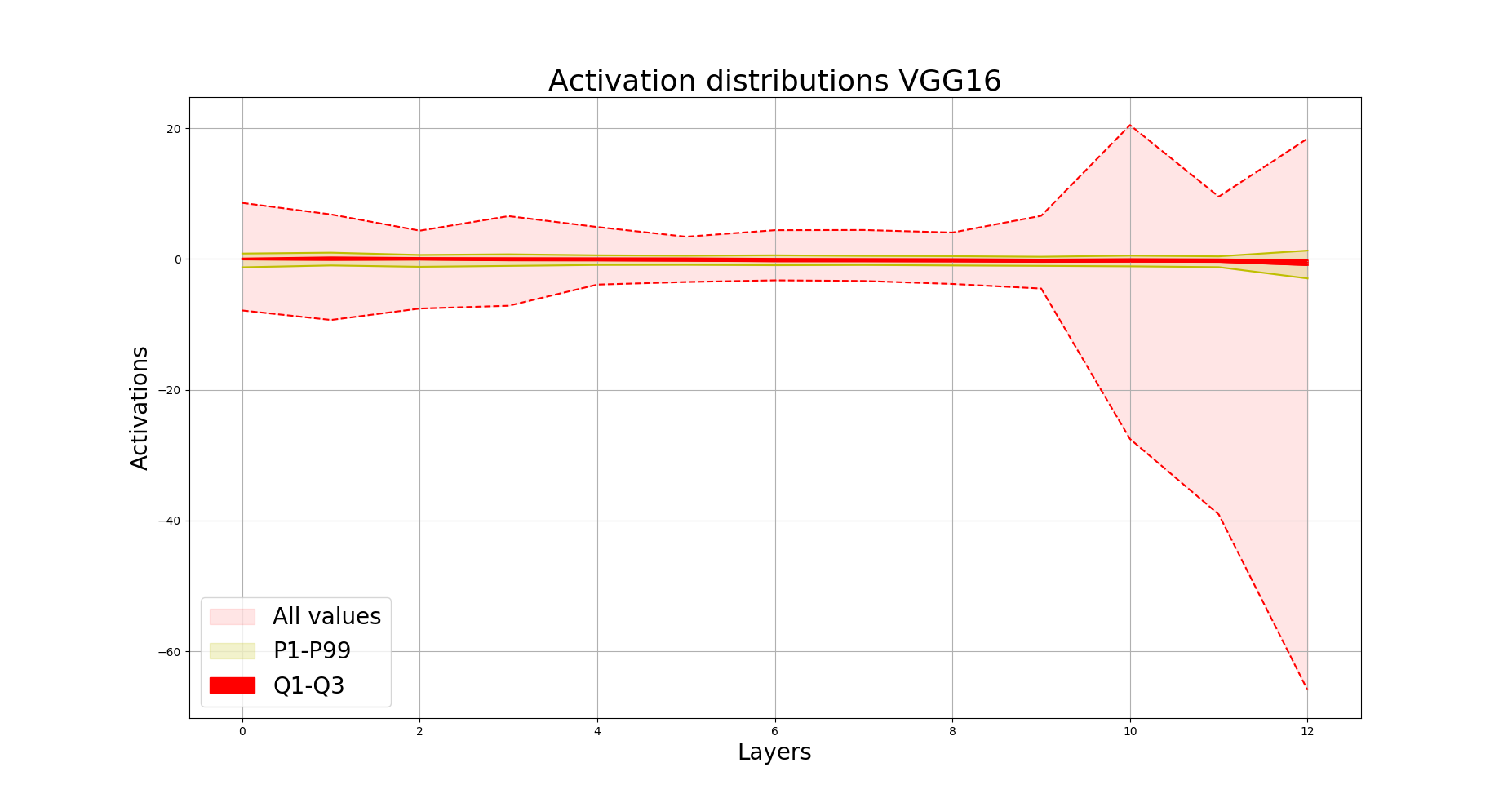}}
    \caption{This figure depicts the distribution of the activations in VGG-19 after the first convolutional layers. As it can be seen the distribution is similar as in case of VGG-16. We have to emphasize that on this plot only activations after convolutional layers are present. $98\%$ can be found in the golden band and the middle dark red band contains $75\%$ of the values. The pretrained weights of the model were downloaded from: \url{https://download.pytorch.org/models/vgg19_bn-c79401a0.pth}}
\label{FigactivationsVGG16}
\end{figure}

\begin{figure}[!htp]
\vskip 0.2in
 \centering
    \subfigure{
    \includegraphics[width=\columnwidth]{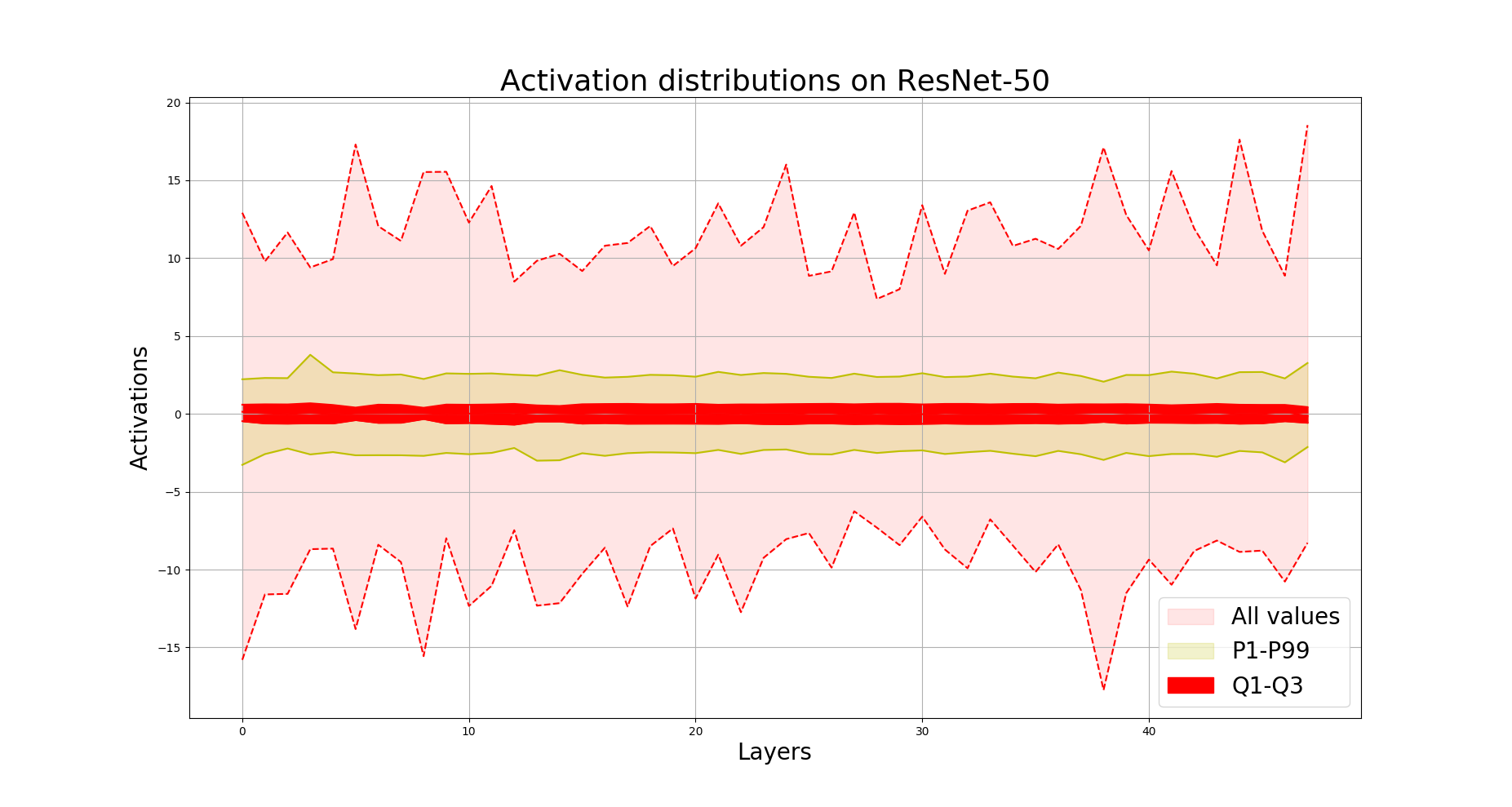}}
    \caption{This figure depicts the distribution of the activations in the ResNet-50 architecture after every batch normalization. As it can be seen the distribution is similar as in case of VGG-16. Although the distribution of the outliers is different (in VGG large activations have appeared at deep layers, here the distribution of the activations does not seem to be dependent on the layer depth. Also the appearance a periodicity can be seen in the minimal/maximal activations, which might depend on order of the convolution in the Bottleneck Block of the residual structure- but wee have not investigated this in detail), it is still true that extremely unlikely activations appear in the network. We have to emphasize that on this plot only activations after convolutional layers are present. $98\%$ can be found in the golden band and the middle dark red band contains $75\%$ of the values. The pretrained weights of the model were downloaded from:
    \url{https://download.pytorch.org/models/resnet50-19c8e357.pth}}
\label{Figactivationsresnet50}
\end{figure}

In Section 2.2 (especially in Figure 1) we have demonstrated that the distribution of the activations in a neural network not always follow Gaussian distribution and the specificity of neurons works against a Gaussian distribution. Here we would like to demonstrate that these extreme outliers can be find in other network architectures as well and they are not specific for the investigated architecture or network weights.
We have investigated the VGG-19 architecture and the ResNet-50 architectures as well and the distribution of the activations of their layers are depicted on Fig. \ref{FigactivationsVGG16}, \ref{Figactivationsresnet50}. As it can be seen on the figures, although the distribution of the outliers are different in all of them it is generally true, that activations are limited in a small interval ($98\%$ of them can be found in the $\pm4\sigma$ band) all of them contain values outside the $\pm15\sigma$ band, which should not happen (has an extremely low likelihood) in case of Gaussian distribution.

Here we would like to emphasize that in VGG architecture extremely large activations have appeared at the end of the network (in the last convolutional and especially in the fully-connected layers), in the ResNet and ResNext architectures these outliers outside $\pm15\sigma$ range were present in the earlier layers as well and appear periodically.
Extreme activations appear at the end of each Bottleneck block, but we did not investigate the reason behind this.
independently from the emergence of these strange periodic patterns the statement of our paper is true for all investigated architectures, that the activations follow Gaussian distribution, but each architecture contain extreme outliers.

We have demonstrated that is untrue that activations follow Gaussian distribution after the transformation of batch normalization. The expectation of these kind of activations from neurons is generally applied in neural network, whether the apply batch normalization or not. Actually this assumption allows the application of batch norm, which can be fully described by a mean and variance value. 

\begin{figure}[!htp]
\vskip 0.2in
 \centering
    \subfigure{
    \includegraphics[width=\columnwidth]{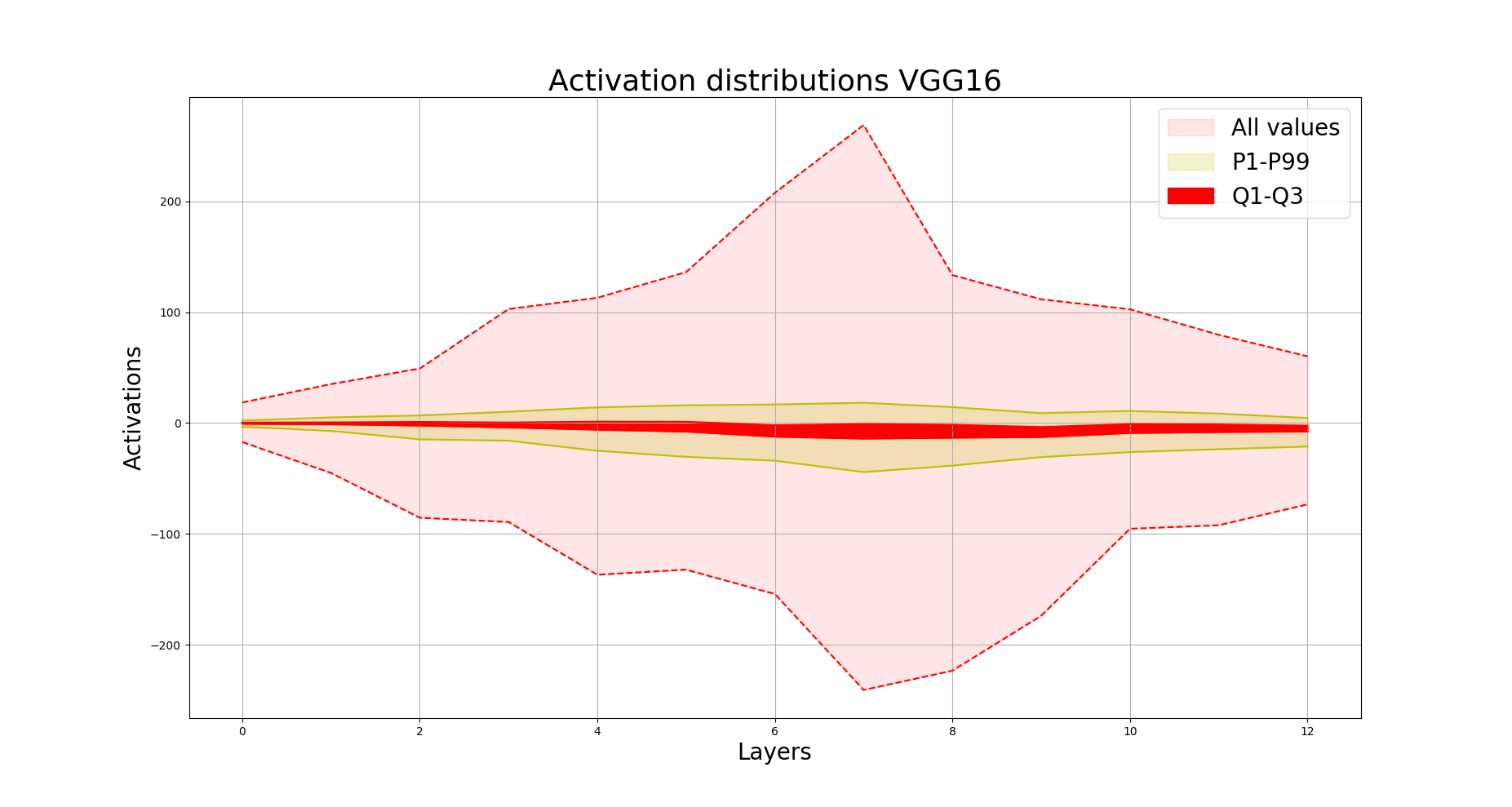}}
    \\
    \subfigure{
    \includegraphics[width=\columnwidth]{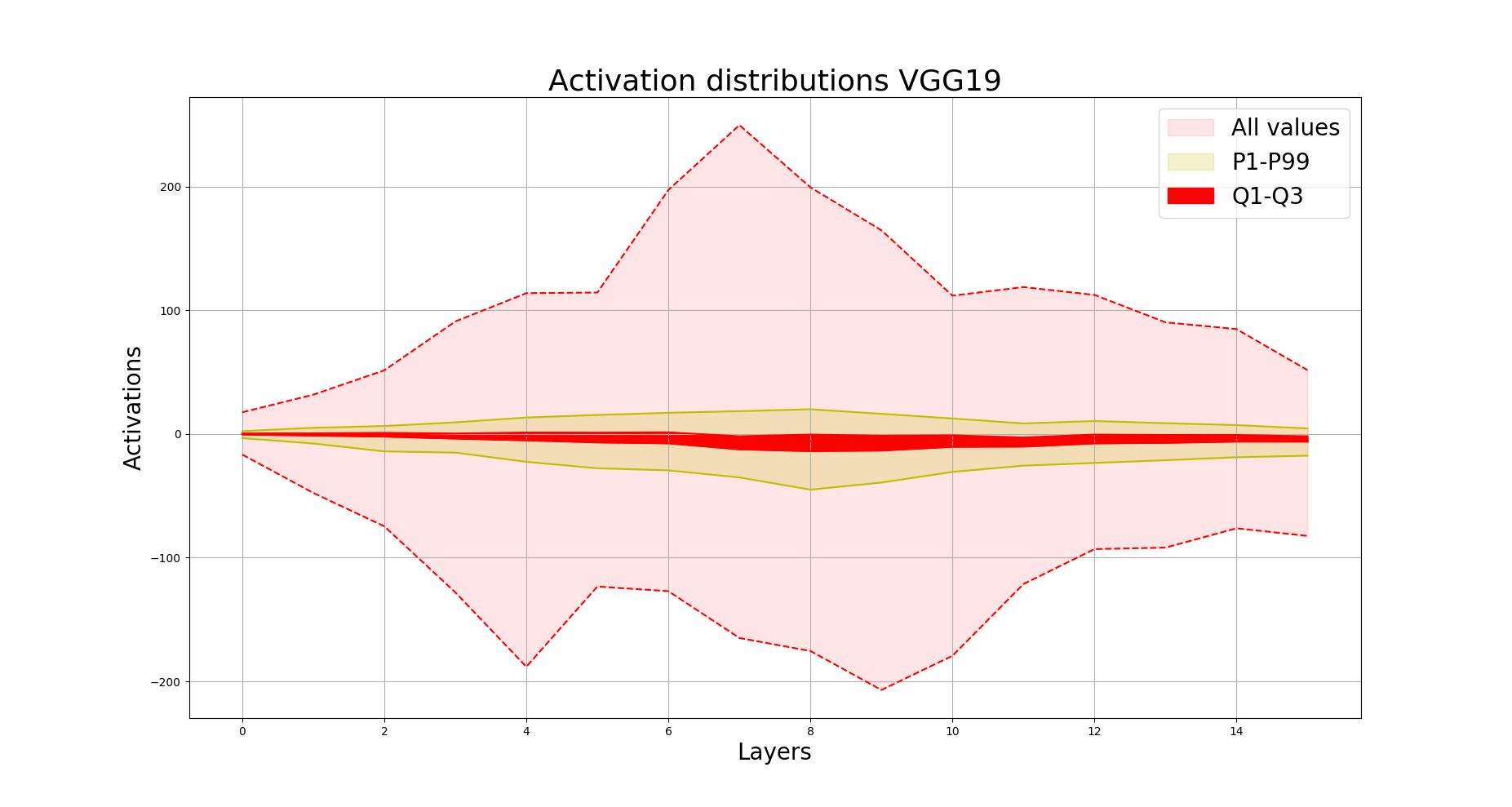}}
    \caption{This figure display the distribution of the activations after each convolutional layers in the VGG-16 and VGG-19 architectures without batch normalization. 
    On this figure $98\%$ of the values are in the gold interval, based on this one can estimate the variance of each distribution (it is not normalized to 1 without batch normalization). one can see, that similarly to batch normalization, extreme activations are present in these cases as well. The pretrained weights were taken from \url{
    https://download.pytorch.org/models/vgg16-397923af.pth} and 
    \url{https://download.pytorch.org/models/vgg19-dcbb9e9d.pth}}
\label{FigactivationsNoBN}
\end{figure}

Because of this we have decided to check the distribution of the activations in the VGG-16 and VGG-19 architecture without batch normalization as well. The distribution of the activations after the convolutional layers can be seen on Fig. \ref{FigactivationsNoBN}.
Here it is different, compared to our previous findings, that large activations appear in the middle of the network, meanwhile large activations appeared mostly around the end of the network with VGG applying batch normalization. But even in this case, without applying batch normalization, the problem motivating filtered batch normalization still exist: that although the majority of the activations follow Gaussian-like distribution, extreme outliers are present.

\section{Selectiveness of Neurons}

In section 2.2 of the original paper (especially in Fig 2.) we have demonstrated that extremely large activation are related to  the specificity of the neurons. We have showed that all inputs, which have caused extremely large activations contained similar features and belonged to the same class.
Here we would like to provide further examples to show that this specificity is not caused only by the sheer lack of kernel selection.

We have selected additional kernels randomly in the  VGG-19 architecture with batch normalisation (an other architecture which was not investigated in the original paper) in the last convolutional layer and investigated those samples which caused a extreme activations outside the $\pm14\sigma$ band.

\begin{figure}[!htp]
\vskip 0.2in
 \centering
    \subfigure{
    \includegraphics[width=0.8in]{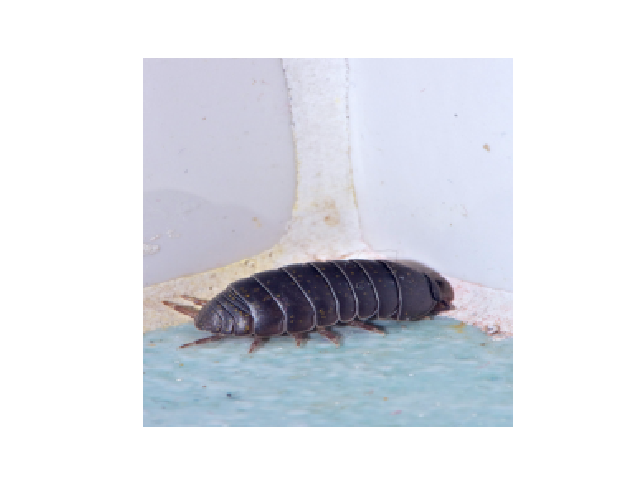}
    \includegraphics[width=0.8in]{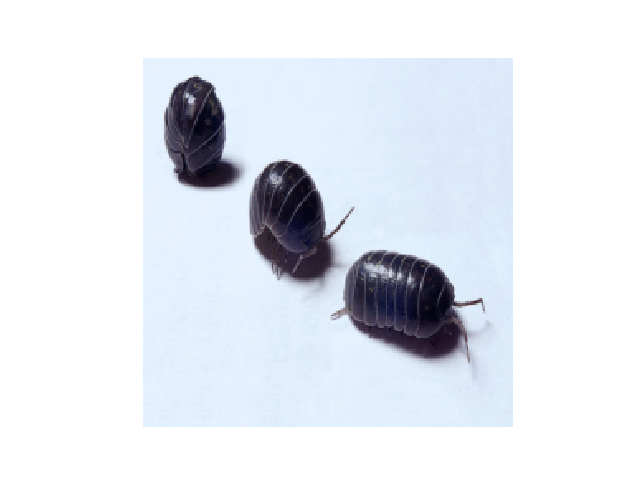}
    \includegraphics[width=0.8in]{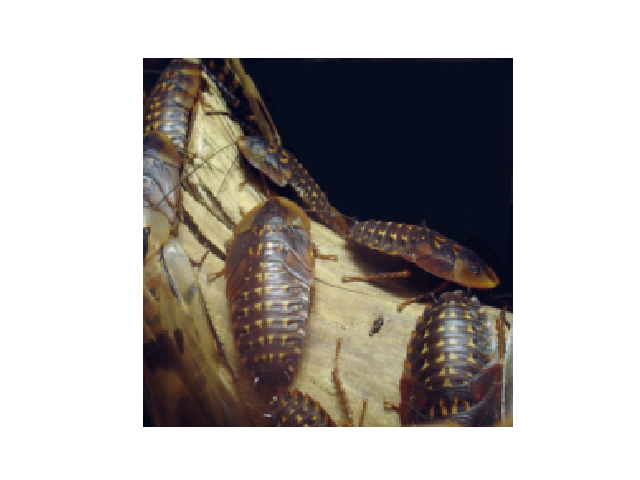}
    \includegraphics[width=0.8in]{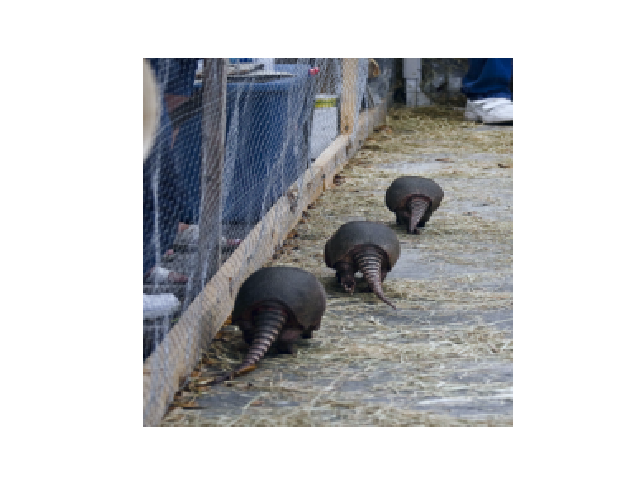}
    }
    \\
    \subfigure{
    \includegraphics[width=0.8in]{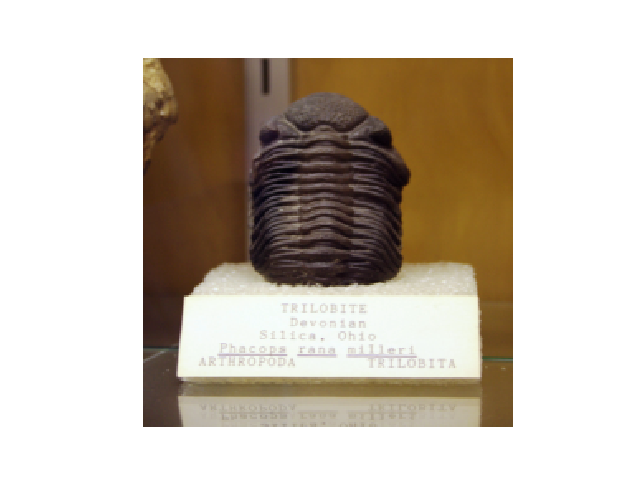}
    \includegraphics[width=0.8in]{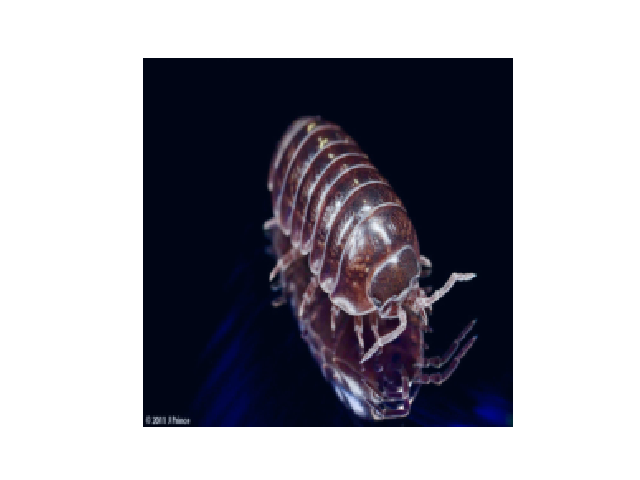}
    \includegraphics[width=0.8in]{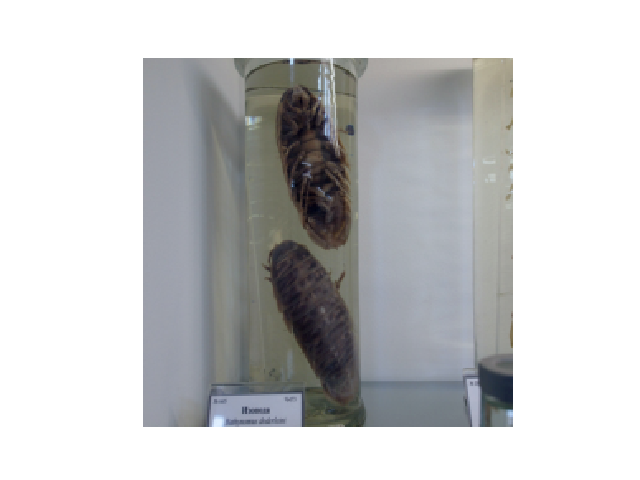}
    \includegraphics[width=0.8in]{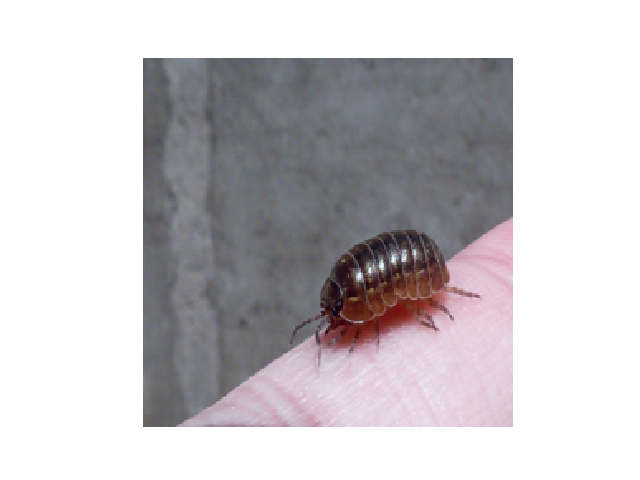}
    }
    \caption{This figure displays randomly selected samples which have caused larger than 14 activations in the 34th kernel of the last convolutional layer of VGG-19. As one can see the images all contain similar features, they contain mostly images of trilobites and isopods. Apart from the image in the top right corner which contain armadillos from the back, which also contain similar features.}
\label{Fig34examplesN}
\end{figure}

\begin{figure}[!htp]
\vskip 0.2in
 \centering
    \subfigure{
    \includegraphics[width=0.8in]{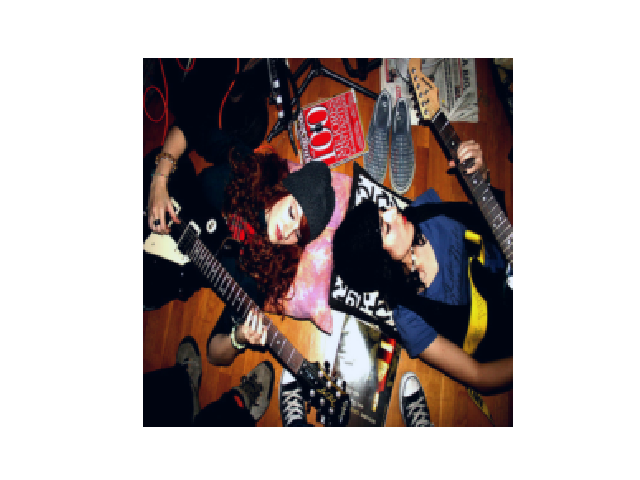}
    \includegraphics[width=0.8in]{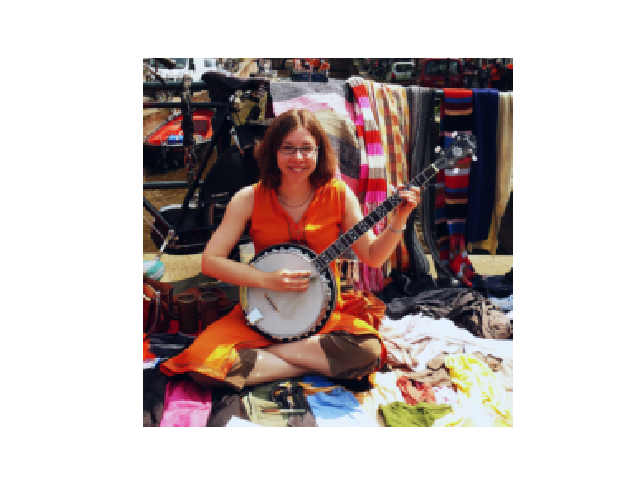}
    \includegraphics[width=0.8in]{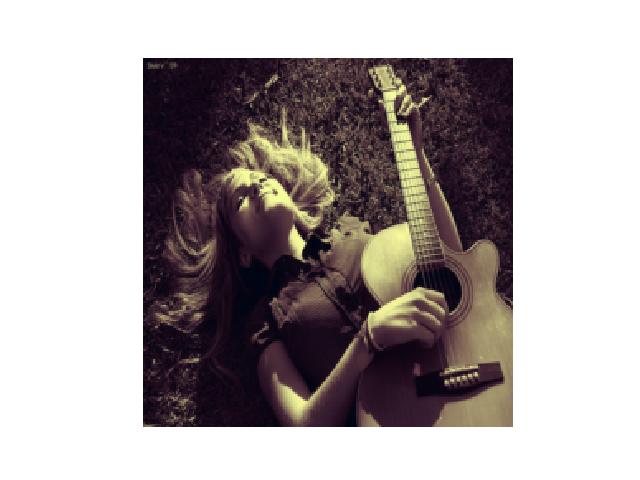}
    \includegraphics[width=0.8in]{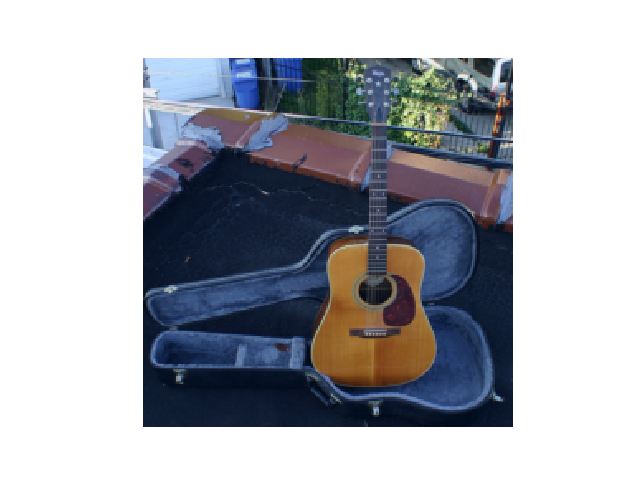}
    }
    \\
    \subfigure{
    \includegraphics[width=0.8in]{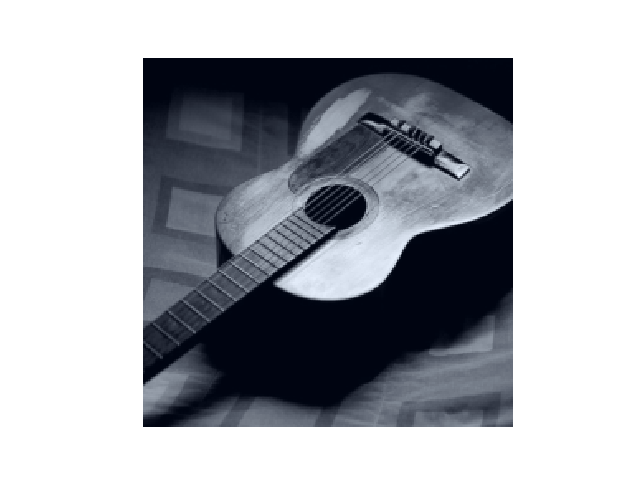}
    \includegraphics[width=0.8in]{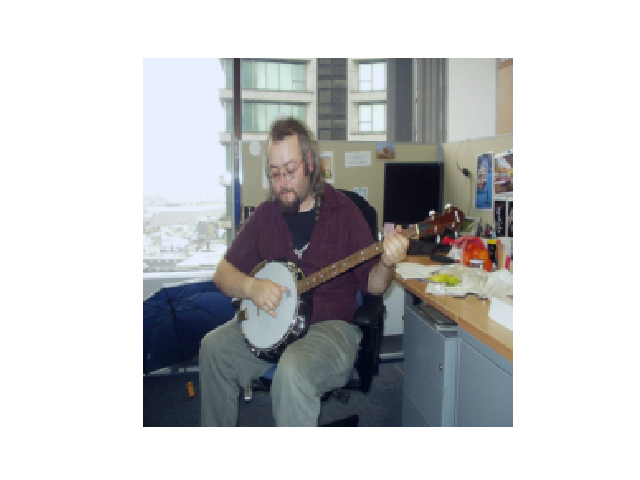}
    \includegraphics[width=0.8in]{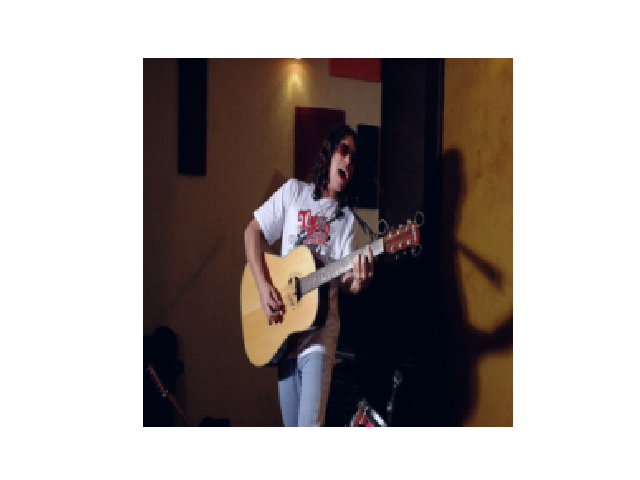}
    \includegraphics[width=0.8in]{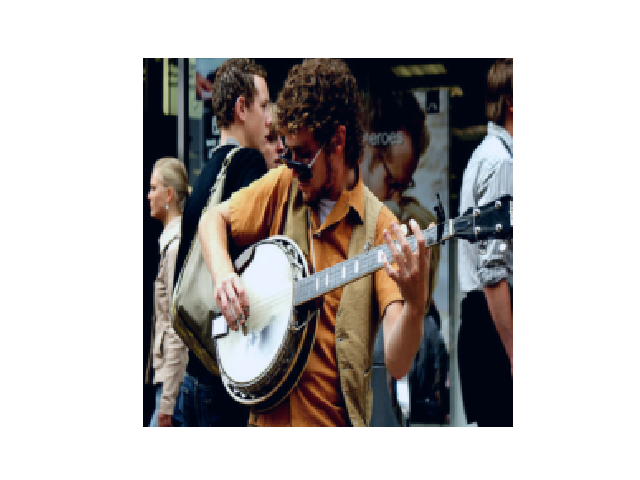}
    }
    \caption{This figure displays randomly selected samples which have caused larger than 14 activations in the 81th kernel of the last convolutional layer of VGG-19. As one can see the images all contain similar features, they are all contain guitars, banjos or other types of plucked string instruments. }
\label{Fig81examplesN}
\end{figure}

\begin{figure}[!htp]
\vskip 0.2in
 \centering
    \subfigure{
    \includegraphics[width=0.8in]{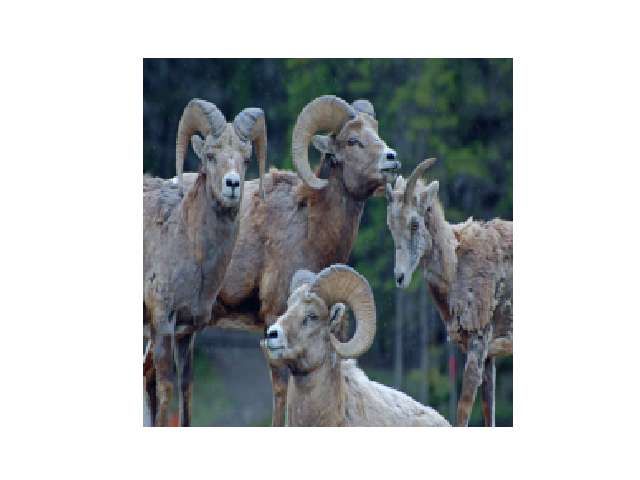}
    \includegraphics[width=0.8in]{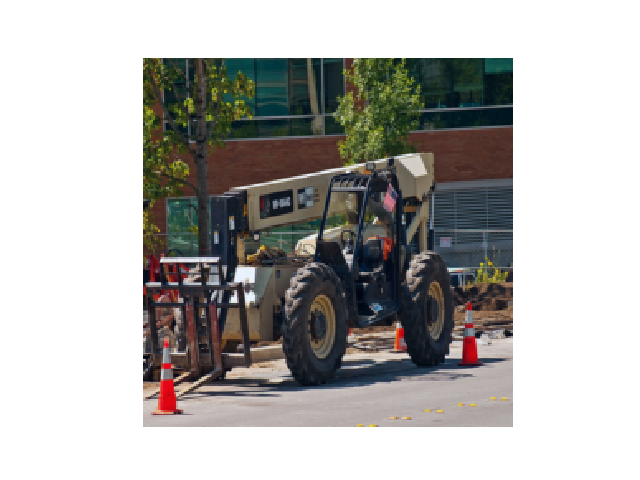}
    \includegraphics[width=0.8in]{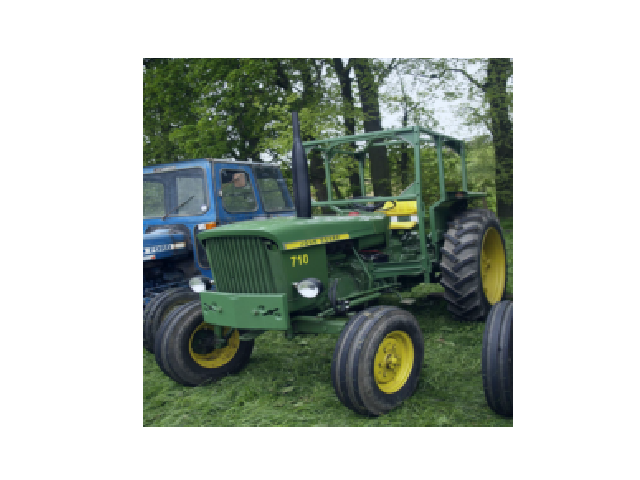}
    \includegraphics[width=0.8in]{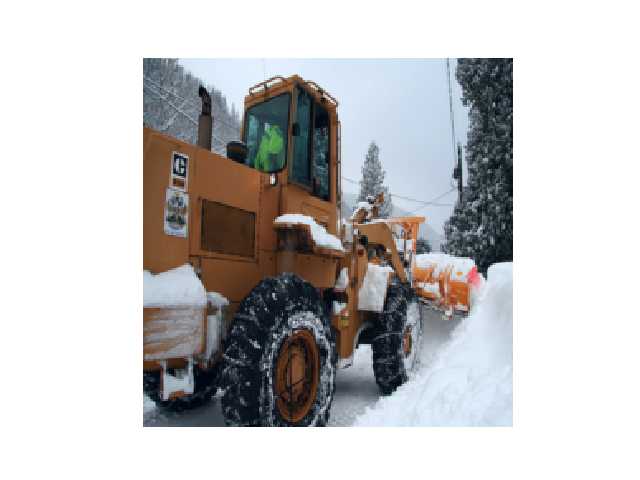}
    }
    \\
    \subfigure{
    \includegraphics[width=0.8in]{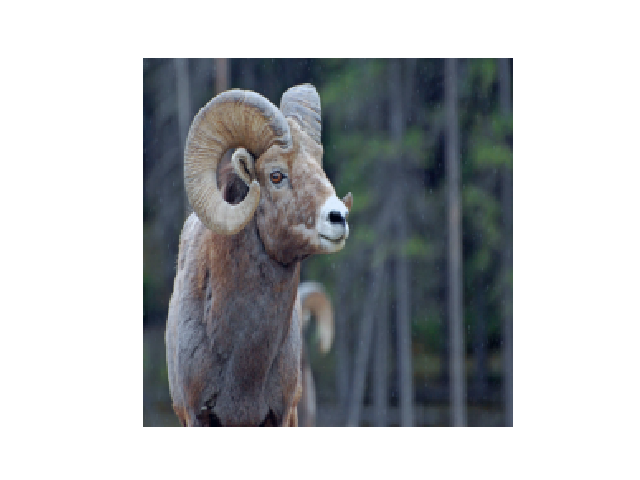}
    \includegraphics[width=0.8in]{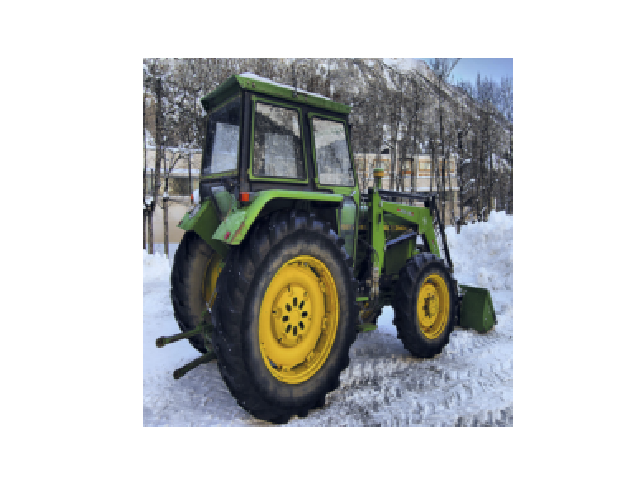}
    \includegraphics[width=0.8in]{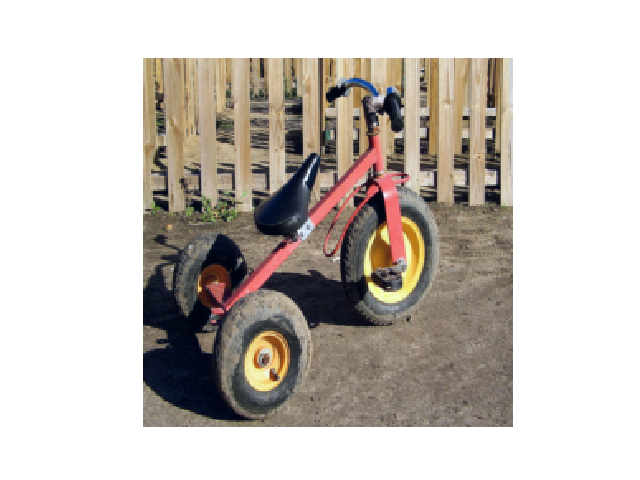}
    \includegraphics[width=0.8in]{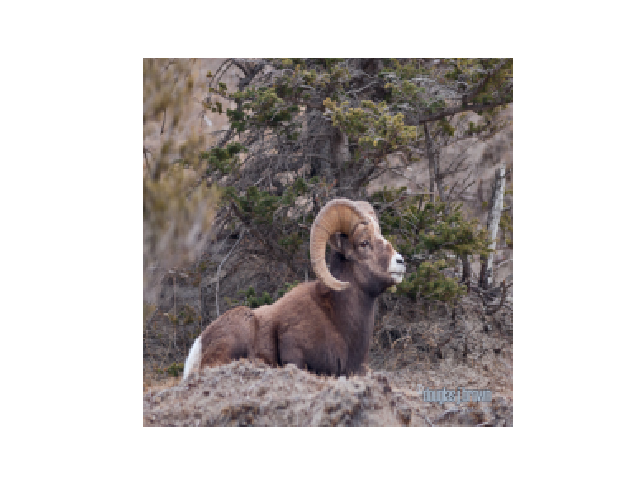}
    }
     \\
    \subfigure{
    \includegraphics[width=0.8in]{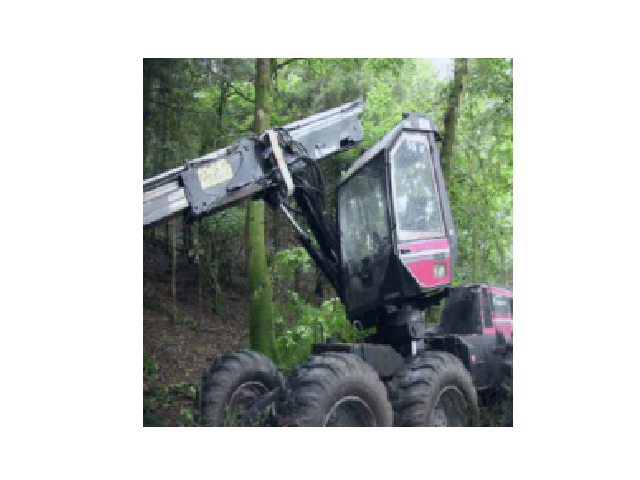}
    \includegraphics[width=0.8in]{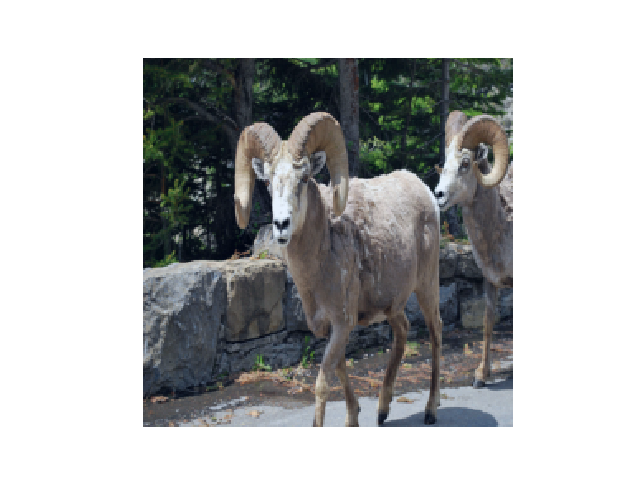}
    \includegraphics[width=0.8in]{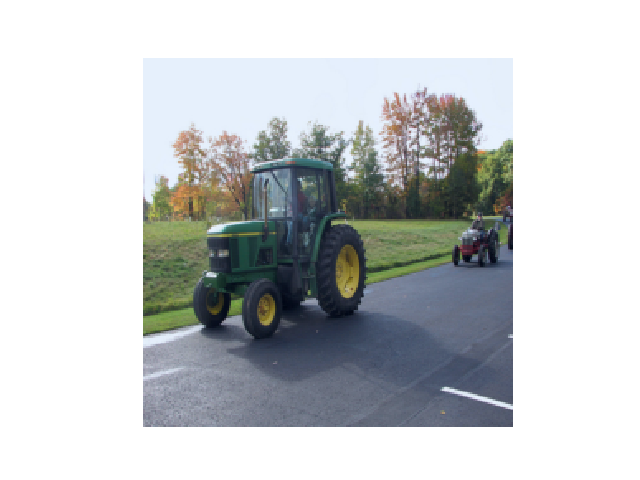}
    \includegraphics[width=0.8in]{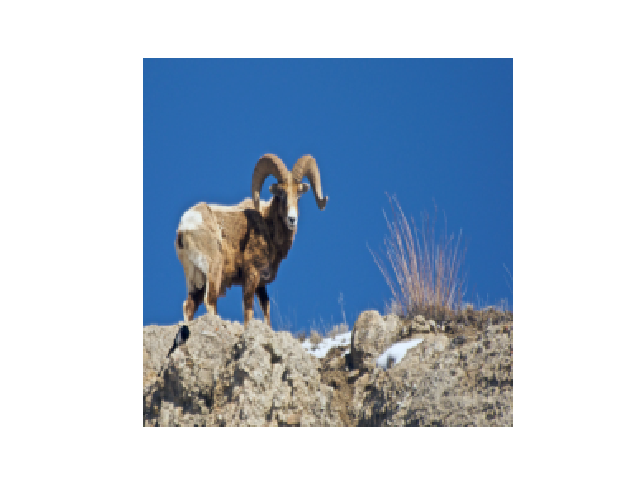}
    }
    \caption{This figure displays randomly selected samples which have caused larger than 14 activations in the 206th kernel of the last convolutional layer of VGG-19. We were not able to associate a single feature with these images, but one can see they can easily be categorized into two categories. All images which invoked an extremely large activation contained either mouflons are vehicles with large wheels (mostly agricultural tractors). These images from ImageNet still are fairly specific and demonstrate that these extreme outliers belong to specific images.}
\label{Fig206examplesN}
\end{figure}

These examples are depicted in Fig. \ref{Fig34examplesN} for the 34th convolution kernel in this layer and \ref{Fig81examplesN} for the 81th and  \ref{Fig206examplesN} for the 206th kernel.

\section{The Layer Dependence of Batch Normalization}

\begin{figure}[!htp]
\vskip 0.2in
 \centering
    \subfigure{
    \includegraphics[width=\columnwidth]{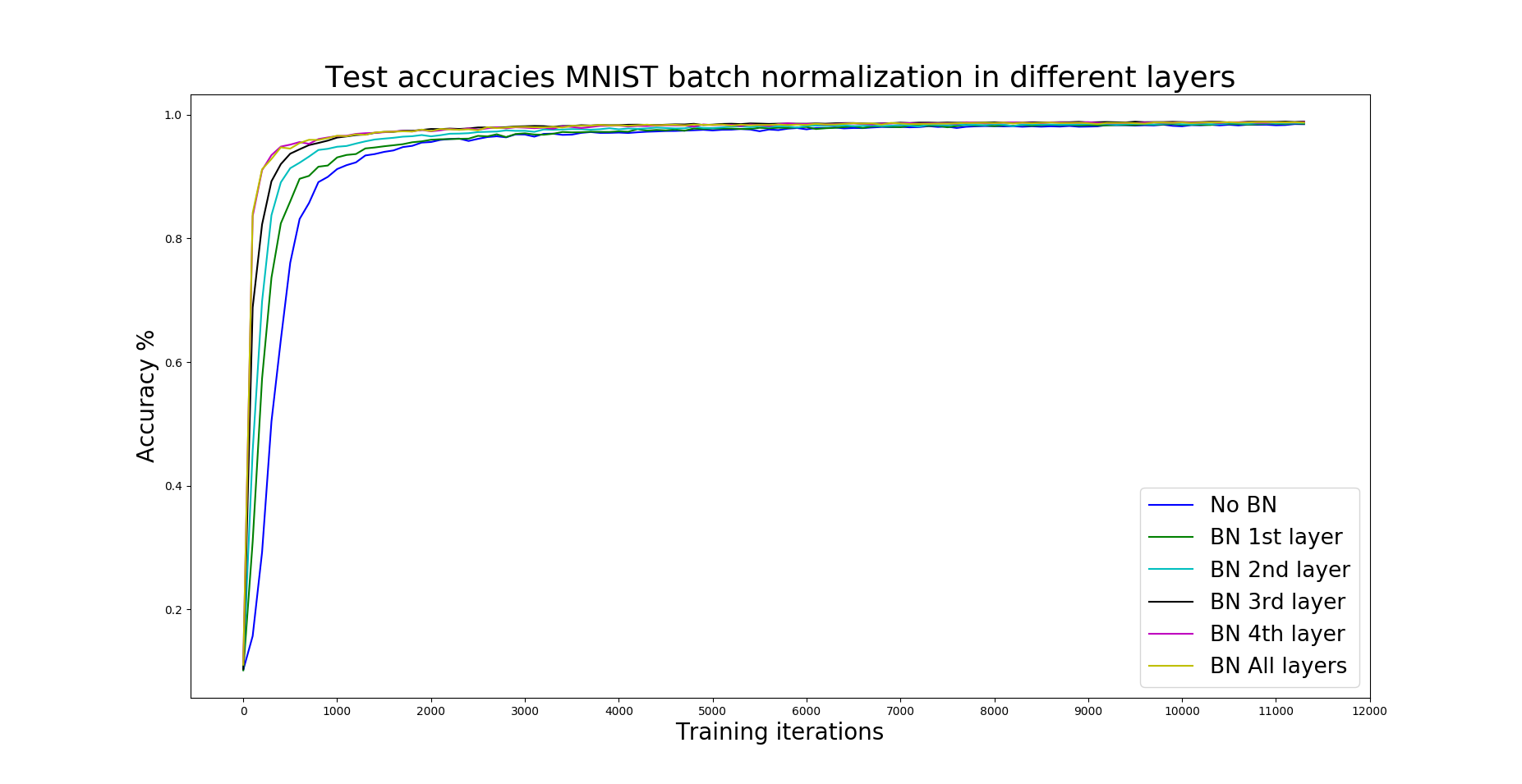}}
    \caption{Test accuracies on the MNIST dataset using the LeNet-5 architecture trained with Adam optimizer with initial learning rate of 0.01 and batches of 32. Batch normalization was applied individually after each layer, this results the four plots in the middle. NO BN was executed without the application of batch normalization and BN ALL layers contained batch normalization after every layer (except the logit layer). As it can be seen from the results batch normalization had the most significant effect after the last feature layer.  Every result displays the average accuracy of ten independent runs.}
\label{FigBNLayers}
\end{figure}

\begin{figure}[!htp]
\vskip 0.2in
 \centering
    \subfigure{
    \includegraphics[width=\columnwidth]{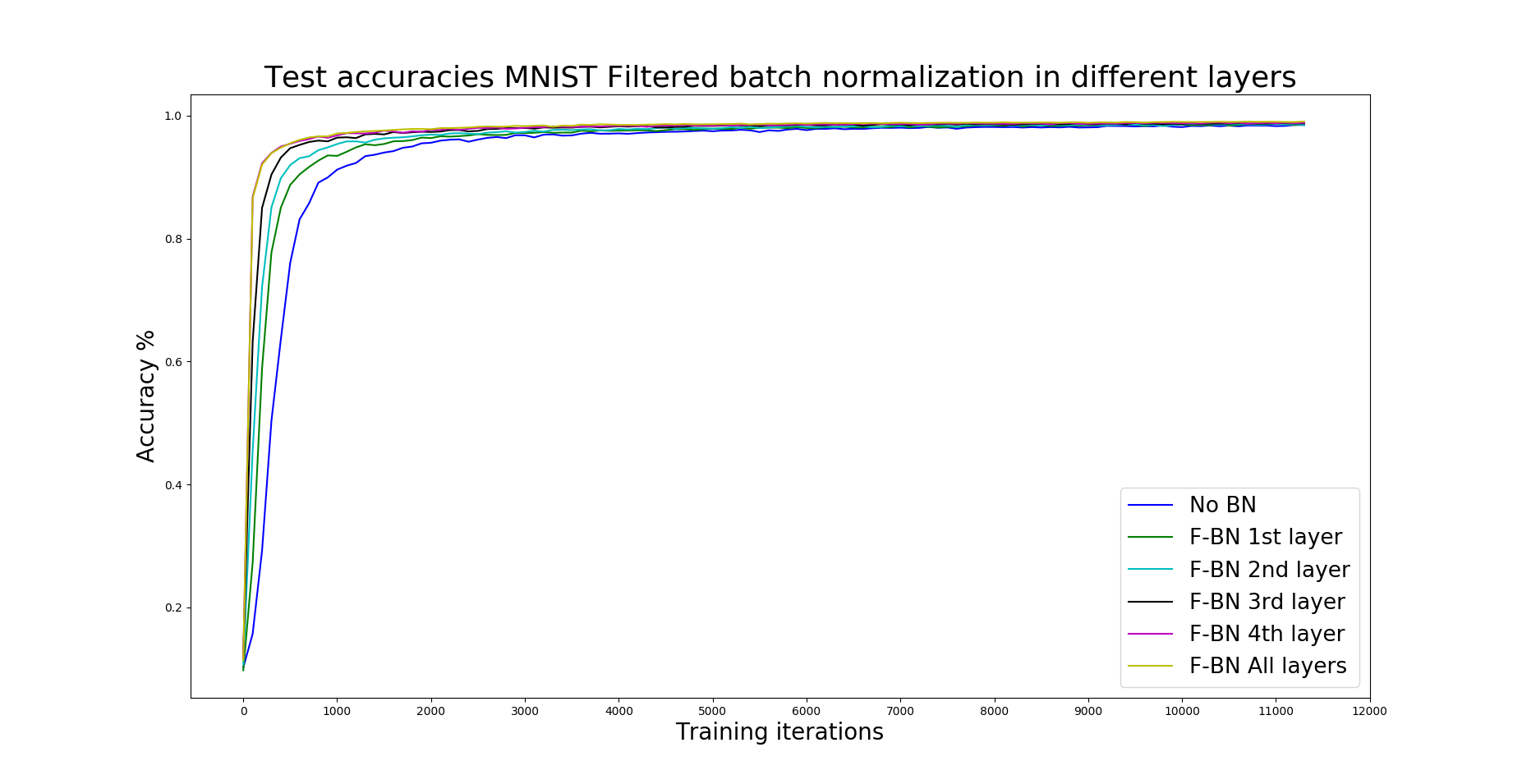}}
    \caption{Test accuracies on the MNIST dataset using the LeNet-5 architecture trained with Adam optimizer with initial learning rate of 0.01 and batches of 32. Fitlered batch normalization was applied individually after each layer, this results the four plots in the middle. NO BN was executed without the application of batch normalization and F-BN ALL layers contained batch normalization after every layer (except the logit layer). As it can be seen from the results filtered batch normalization had the most significant effect after the last feature layer. Every result displays the average accuracy of ten independent runs.}
\label{FigFBNLayers}
\end{figure}

\begin{figure}[!htp]
\vskip 0.2in
 \centering
    \subfigure{
    \includegraphics[width=\columnwidth]{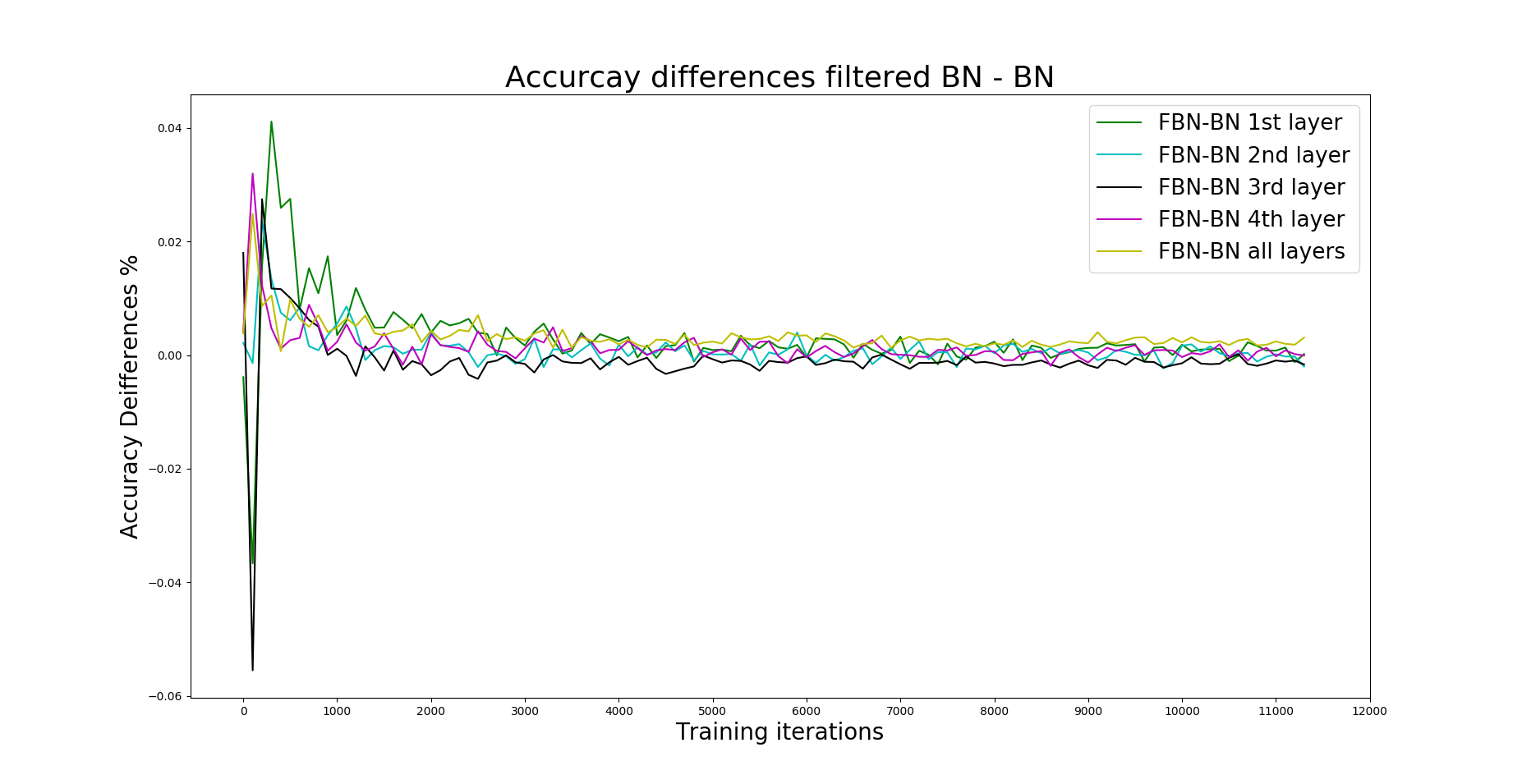}}
    \caption{This figure depicts the differenceso f test accuracies between regular and filtered batch normalization on the MNIST dataset using the LeNet-5 architecture trained with Adam optimizer with initial learning rate of 0.01 and batches of 32. The normalization method was applied after each layer and these results are depicted with different colors. Although the results are fairly noisy it can be seen that the filtered batch normalization caused the largest difference (compared to regular batch normalization) when it was applied in the first or the last layer.}
\label{FigBNFBNDIFF}
\end{figure}

According to the hypothesis introduced in \cite{ioffe2015batch}, that batch normalization helps to reduce internal covariate shift, its effect should be the most significant, if it is applied in the first layers of a neural network, since these layers are trained in the slowest manner. To investigate this we have investigated the effect of batch normalization on the MNIST dataset using the LeNet-5 architecture, which originally does not contain batch normalization and checked how test accuracies improve if we add batch normalization separately after each layer.
The results can be seen in Fig \ref{FigBNLayers} for regular batch normalization and in Fig. \ref{FigFBNLayers} for filtered batch normalization. As it can be seen from the results batch normalization results the highest test accuracy if it is applied after the last feature layer (right before the logit layer). This goes against the hypothesis of \cite{ioffe2015batch} and underline the findings of \cite{santurkar2018does} that internal covariate shift plays only a minor role and the most important benefits of batch normalization is that it creates a smoother loss and gradient landscape. Loss changes depend most on the operations of the last layer this way it underlines our findings.
As it can be seen from these two figures batch normalization and filtered batch normalization seems to have a similar effect regarding after which layer it is applied. Naturally its application does not impair test accuracy, so for the best result can be reach if it is applied after every layer, but similar accuracy was reached when it was only applied right before the logit layer.

we wanted to investigate the effect of batch normalization and filtered batch normalization in details and because of this we have also plotted the difference between the test accuracies of regular batch normalization and filtered batch normalization for the average of ten independent runs. The results are displayed in Fig. \ref{FigBNFBNDIFF}. Although the results are fairly noisy, one can observe that the filtered batch normalization have larger and larger effect in the deeper and deeper layers (unfortunately the first layer is an exception, where the difference is the most significant). We would like to think that this is caused by the fact that more and more extreme outliers appear in later and later layers, which change the mean and variance values of batch normalization more and more drastically and where filtered batch norm can help more. But we have to admit that these results (in Fig. \ref{FigBNFBNDIFF}) are noisy, difficult to be interpreted, but most importantly they are only preliminary and require more and detailed investigation.

\bibliographystyle{IEEEtran}
\bibliography{batchnorm.bib}




